\documentclass[multiauthors,onecolumn,cfinal]{LaTeXclass/cohere}

\usepackage{booktabs}
\usepackage{colortbl}
\usepackage{graphicx}
\usepackage{hyperref}
\usepackage{caption}
\usepackage{geometry}
\usepackage{CJKutf8}
\usepackage{multirow}
\usepackage{makecell}
\usepackage{textcomp}
\usepackage{enumitem}
\usepackage{ragged2e}
\usepackage{soul}
\usepackage{wrapfig}

\usepackage{arydshln}
\usepackage{tikz}

\usepackage{fontawesome}
\usepackage{xspace}
\newcommand{\huggingface}{\raisebox{-1.5pt}{\includegraphics[height=1.05em]{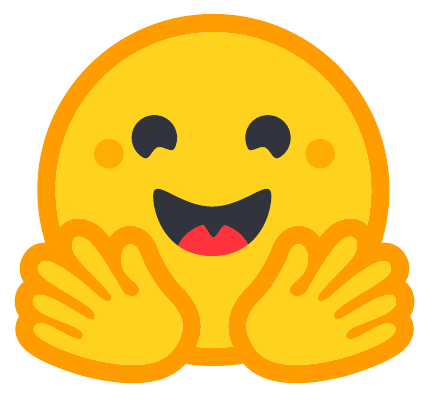}}\xspace}
\newcommand{\github}{\raisebox{-1.5pt}{\includegraphics[height=1.05em]{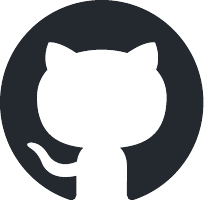}}\xspace}
\newcommand{\worldwideweb}{\raisebox{-1.5pt}{\includegraphics[height=1.05em]{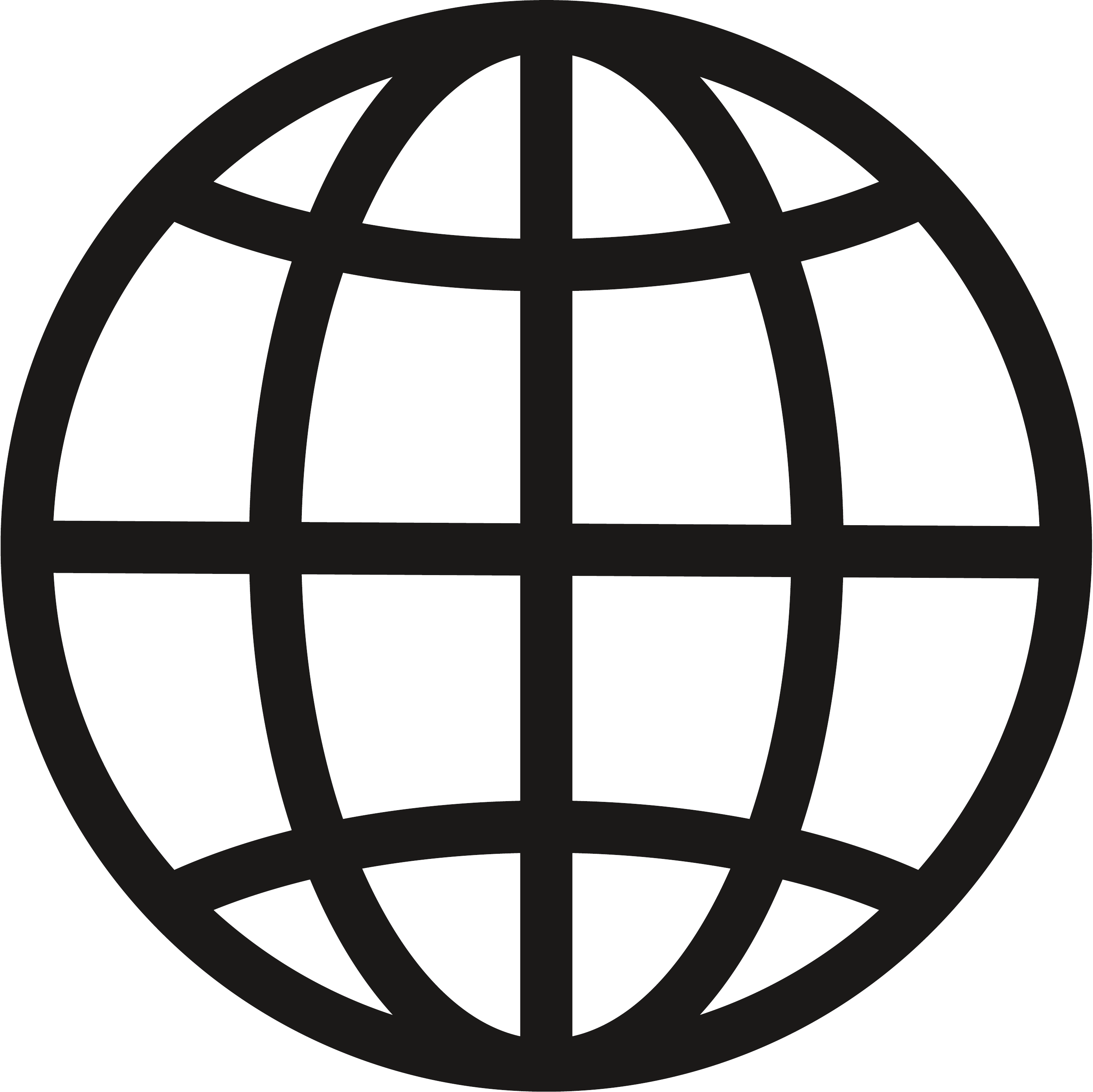}}\xspace}

\definecolor{lightblue}{HTML}{DDEEFF}
\definecolor{lightgreen}{HTML}{DFF0D8}
\definecolor{lightyellow}{HTML}{FFF9DD}
\definecolor{lightorange}{HTML}{FFE5CC}

\newcommand{\ours}{\textsc{NeoBabel}\xspace}

\setcitestyle{number,square}

\title{
\begin{minipage}{0.16\textwidth}
    \includegraphics[scale=0.12]{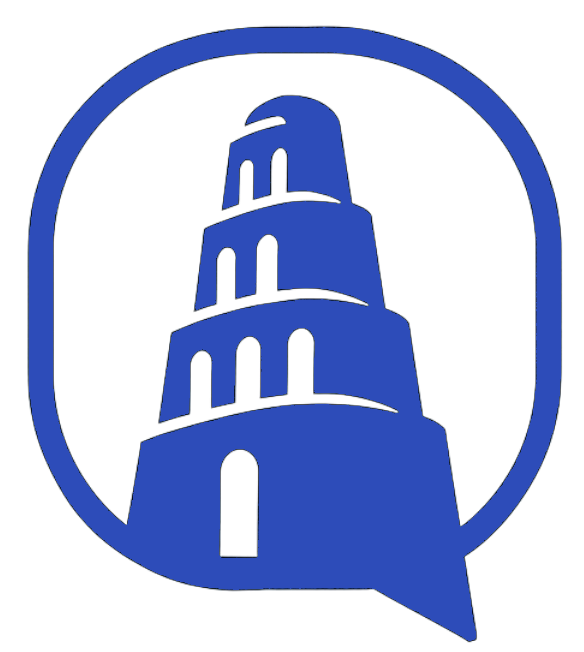}
\end{minipage}
\hspace{-0.4cm}
\begin{minipage}{0.80\textwidth}
    {NeoBabel: A Multilingual Open Tower for Visual Generation}
\end{minipage}
}
\author{name={Mohammad Mahdi Derakhshani},affiliation={2}}
\author{name={Dheeraj Varghese},affiliation={2}}
\author{name={Marzieh Fadaee\psa},affiliation={1}}
\author{name={Cees G. M. Snoek\psa},affiliation={2}}
\affiliations{
    \item[1] Cohere Labs
    \item[2] University of Amsterdam
}

\corresponding[*]{\url{m.m.derakhshani}\url{{@uva.nl}}, \url{marzieh}\url{{@cohere.com}}}

\abstract{
\justifying
Text-to-image generation advancements have been predominantly English-centric, creating barriers for non-English speakers and perpetuating digital inequities. While existing systems rely on translation pipelines, these introduce semantic drift, computational overhead, and cultural misalignment. We introduce \ours, a novel multilingual image generation framework that sets a new Pareto frontier in performance, efficiency and inclusivity, supporting six languages: \textit{English, Chinese, Dutch, French, Hindi}, and \textit{Persian}. The model is trained using a combination of large-scale multilingual pretraining and high-resolution instruction tuning. To evaluate its capabilities, we expand two English-only benchmarks to multilingual equivalents: m-GenEval and m-DPG. \ours achieves state-of-the-art multilingual performance while retaining strong English capability, scoring 0.75 on m-GenEval and 0.68 on m-DPG. 
Notably, it performs on par with leading models on English tasks while outperforming them by +0.11 and +0.09 on multilingual benchmarks, even though these models are built on multilingual base LLMs.
This demonstrates the effectiveness of our targeted alignment training for preserving and extending cross-lingual generalization.
We further introduce two new metrics to rigorously assess multilingual alignment and robustness to code-mixed prompts. Notably, \ours matches or exceeds English-only models while being 2–4× smaller. We release an open toolkit, including all code, model checkpoints, a curated dataset of 124M multilingual text-image pairs, and standardized multilingual evaluation protocols, to advance inclusive AI research. Our work demonstrates that multilingual capability is not a trade-off but a catalyst for improved robustness, efficiency, and cultural fidelity in generative AI. 
}

\begin{document}

% Inspiration: https://arxiv.org/abs/2402.00838
\begin{center}
    \renewcommand{\arraystretch}{1}
    \begin{tabular}{rll}
        \worldwideweb & \textbf{Website} & \url{https://Neo-Babel.github.io}\\
        \github & \textbf{Code} & \url{https://github.com/mmderakhshani/NeoBabel}\\
        \huggingface & \textbf{Models} & \url{https://hf.co/mderakhshani/NeoBabel}\\
        \huggingface & \textbf{Pretraining Data} & \url{https://hf.co/datasets/mderakhshani/NeoBabel-Pretrain}\\
        \huggingface & \textbf{Instruction Data} & \url{https://hf.co/datasets/mderakhshani/NeoBabel-Instruct}\\
        \huggingface & \textbf{Evaluation Data} & \url{https://hf.co/datasets/mderakhshani/NeoBabel-Eval}\\
    \end{tabular}
\end{center}

\begin{figure}[htb]
    \centering
    \includegraphics[width=1\linewidth, trim=0 0 10 0, clip]{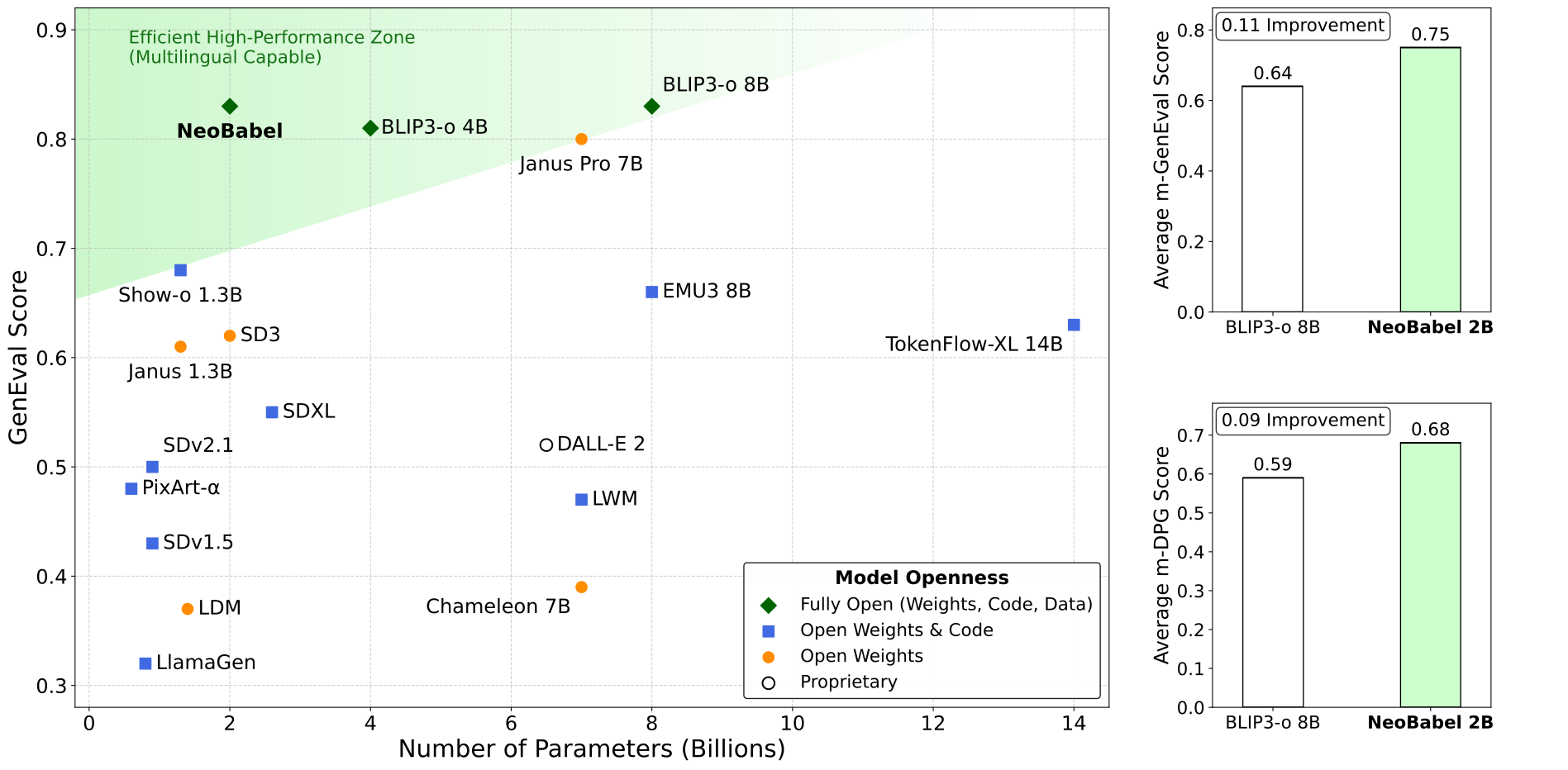}
    \caption{\textbf{NeoBabel establishes a new Pareto frontier in multilingual image generation performance, efficiency, and inclusivity.} Left: GenEval English-only scores show that \ours matches state-of-the-art models despite being 2–4× smaller. Right: On our multilingual benchmark extensions, m-GenEval and m-DPG, \ours outperforms the second-best model, demonstrating strong multilingual generalization. \ours is fully open (weights, code, data) and supports six languages with consistent cross-lingual performance.}
    \label{fig:geneval_demo}
\end{figure}

\section{Introduction}
\label{sec:intro}

Recent advances in diffusion models and large-scale vision-language pretraining have revolutionized text-to-image generation, enabling the creation of high-quality images from natural language descriptions \citep{rombach2022high,peebles2023scalable,uvit,pixart,Xie_2023_ICCV,wu2023tune,lipman2022flow,xie2025sana,qin2025lumina,show1,seawead2025seaweed}. Despite these remarkable capabilities, the field suffers from a critical limitation: an overwhelming reliance on English as the primary—and often exclusive—input language \citep{ramesh2022hierarchical,xie2024show,team2024chameleon}. This monolingual bias creates substantial barriers for the billions of users who communicate in other languages, fundamentally restricting global access to state-of-the-art generative AI technologies \citep{bassignana2025theaigap,peppin2025multilingualdivideimpactglobal}.
The consequences of this linguistic limitation extend far beyond mere inconvenience. As text-to-image systems become integral to education, creative industries, art, and journalism, the lack of native multilingual support perpetuates existing digital divides and cultural inequities \citep{liu2023onthecultural,rege2025cureculturalgaps}. 
Non-English speakers are forced to navigate through translation layers that not only introduce friction but also risk losing the nuanced meanings and cultural contexts that make their creative expressions unique \citep{kannen2024beyondaestheticscultural,friedrich2024multilingualtextto}. 
Building truly multilingual models, like we do in this paper, is therefore not merely a technical challenge but an ethical imperative, one that ensures equitable access to generative AI while preserving linguistic diversity and cultural authenticity in the digital age.

Existing approaches to multilingual image generation typically employ a translation-first strategy, converting non-English prompts to English before processing. 
While this appears pragmatic, it introduces a cascade of problems that fundamentally compromise the user experience \citep{kreutzer2025dejavumultilingualllm,li2025lostliteralismsupervisedtraining,bafna2025translation}. 
The computational overhead of chaining translation and generation models effectively doubles inference time, creating prohibitive delays for real-time applications, thereby further disadvantaging non-English speakers. Most critically, this approach suffers from semantic drift—the systematic loss of culturally specific meanings and linguistic subtleties \citep{cohn-gordon-goodman-2019-lost,vanmassenhove-etal-2019-lost,beinborn-choenni-2020-semantic}. 
For instance consider the Dutch term ``\textit{gezellig}'' which encompasses a complex blend of coziness, conviviality, and belonging and has no direct English equivalent. When forced through translation, such rich cultural concepts are inevitably flattened or distorted, resulting in generated images that fail to capture the intended meaning. The fundamental issue lies deeper than mere translation accuracy \citep{wein2023lost,singh2024global,salazar2025kaleidoscope}. 

Current vision-language architectures treat multilingual support as an afterthought, forcing diverse linguistic communities to conform to English-centric models rather than developing systems that natively understand and respect linguistic diversity. This design philosophy not only limits accessibility but also wastes the potential benefits of multilingual training, which could enhance model robustness, cross-cultural understanding, and generalization capabilities across different linguistic and cultural contexts \citep{ji2024can,faisal-anastasopoulos-2024-efficient,dash2025ayavisionadvancingfrontier,shimabucoro2025post}.
These challenges demand a paradigm shift toward native multilingual understanding in text-to-image generation. The primary obstacle remains the scarcity of high-quality, culturally annotated visual-linguistic datasets for non-English languages. 
Even with adequate data, significant technical barriers persist: establishing robust cross-lingual concept alignment, modeling typological variations across language families, and preserving culture-specific semantics during generation. Overcoming these limitations is critical for transitioning from mere translation-based approaches to systems with genuine multilingual competence.

This paper introduces \ours, a novel multilingual image generation framework that represents the first scalable solution for direct text-to-image synthesis across six languages.
Through meticulous curation of high-quality multilingual vision-language datasets and end-to-end training, \ours establishes direct cross-lingual mappings between textual descriptions and visual outputs across all supported languages. This approach not only removes translation dependencies but also maintains crucial cultural and linguistic specificity in the generated images.
Our model demonstrates that multilingual capability isn't a trade-off but rather a catalyst for improved model performance. 

Our work addresses three key questions: 1) \textit{How can we train a single model to handle multiple languages effectively?} 2) \textit{Does multilingual training degrade performance in high-resource languages like English?} and 3) \textit{Can a unified model outperform language-specific or translation-based approaches?}
To answer these, we introduce a progressive training pipeline that combines large-scale multilingual pretraining with high-resolution instruction tuning. We evaluate \ours on m-GenEval and m-DPG, our multilingual extensions of GenEval~\citep{ghosh2023geneval} and DPG-Bench~\citep{hu2024ella}, and introduce two new metrics, Cross-Lingual Consistency (CLC) and Code Switching Similarity (CSS), to quantify multilingual performance.

As shown in Figure~\ref{fig:geneval_demo}, \ours matches the performance of state-of-the-art English-only models while being 2–4× smaller. Here, we report English-only results for fair comparison, as prior work evaluates only in English. Furthermore, \ours maintains strong generation quality in all six supported languages. For instance, on the m-GenEval benchmark, it achieves a new state-of-the-art score of 0.75—an improvement of 0.11 over the very recent BLIP3-o 8B model (0.64)~\citep{chen2025blip3}. Similarly, on m-DPG, it scores 0.68, outperforming BLIP3-o 8B by 0.09. These results demonstrate that strong multilingual generation is achievable without resorting to large-scale models or sacrificing output quality.

To summarize, we make the following key contributions:
\begin{enumerate}
\item \textbf{A novel multilingual training framework.} We introduce a novel multilingual training framework that establishes new state-of-the-art performance in cross-lingual image generation. Our approach achieves language-agnostic understanding by directly mapping prompts from any supported language to visual concepts without requiring translation, while maintaining performance parity that matches or exceeds English-only models across all languages. This unified architecture delivers significant operational efficiency gains by eliminating the need for separate translation infrastructure, enabling single-model deployment that reduces both computational overhead and system complexity. The unified architecture delivers significant efficiency improvements, processing multilingual prompts 2.8x faster than translation-then-generation pipelines while using 59\% less memory which is critical for real-world deployment scenarios. 
To train the unified model, we introduce a data curation pipeline that prepares multilingual image-text pairs for both pretraining and instruction tuning.
\item \textbf{Comprehensive multilingual benchmark and metrics.} We introduce the first standardized framework for evaluating multilingual image generation, addressing critical gaps in existing benchmarks. Our protocol includes: (1) extended versions of GenEval~\citep{ghosh2023geneval} and DPG-Bench~\citep{hu2024ella}, referred to as m-GenEval and m-DPG, across six languages, enabling direct comparison between native multilingual and translation-based approaches; and (2) two novel metrics—Cross-Lingual Consistency (CLC) and Code-Switching Similarity (CSS), to quantify semantic alignment and robustness to mixed-language prompts (see Figure~\ref{fig:cross-lingual-qualitative}). CLC measures image equivalence across languages using EVA-CLIP~\citep{sun2023eva} and DINOv2~\citep{oquab2023dinov2} embeddings, while CSS evaluates real-world code-switching scenarios. \ours achieves state-of-the-art multilingual performance while maintaining strong English capabilities. Notably, it matches the English results of leading multilingual models while outperforming them by +0.11 and +0.09 on multilingual benchmarks—despite those models being built on multilingual base LLMs. This positions \ours as a strong foundation for future research in equitable, culturally adaptive generative AI.
\item \textbf{Open toolkit for inclusive research.} We release a comprehensive research toolkit comprising \ours model checkpoints trained on six languages (English, Chinese, Dutch, French, Hindi, and Persian), a systematically curated dataset of 124M multilingual text-image pairs with quality-controlled translations, and a complete reproducibility package including training scripts, hyperparameter configurations, and standardized evaluation protocols. Our framework is designed to be easily extensible to additional languages, thanks to a scalable training pipeline, with validation metrics and benchmarking guidelines that support systematic comparison of multilingual generation across research groups.
\end{enumerate}
In the following sections, we present the details of \ours, including its architecture (Section~\ref{sec:neobabl_arch}), multilingual datasets (Section~\ref{sec:neobabel_multilingual_datasets}), progressive training stages (Section~\ref{sec:neobabel_training_stages}), and multilingual evaluation suite (Section~\ref{sec:eval_suite}). 
We then provide both quantitative and qualitative evaluations (Section~\ref{sec:experiment}), followed by ablation studies and analysis (Section~\ref{sec:ablation}). 

\section{NeoBabel Architecture}
\label{sec:neobabl_arch}

We first outline the core architectural components of \ours, including its multilingual transformer backbone (Section~\ref{sec:model-architecture}), training objectives (Section~\ref{sec:training-objective}), and the multilingual model merging strategy (Section~\ref{sec:model-merging-method}) designed to enhance generation quality across diverse linguistic settings.

\subsection{Model Architecture}
\label{sec:model-architecture}
\begin{figure}
    \centering
    \includegraphics[width=\linewidth]{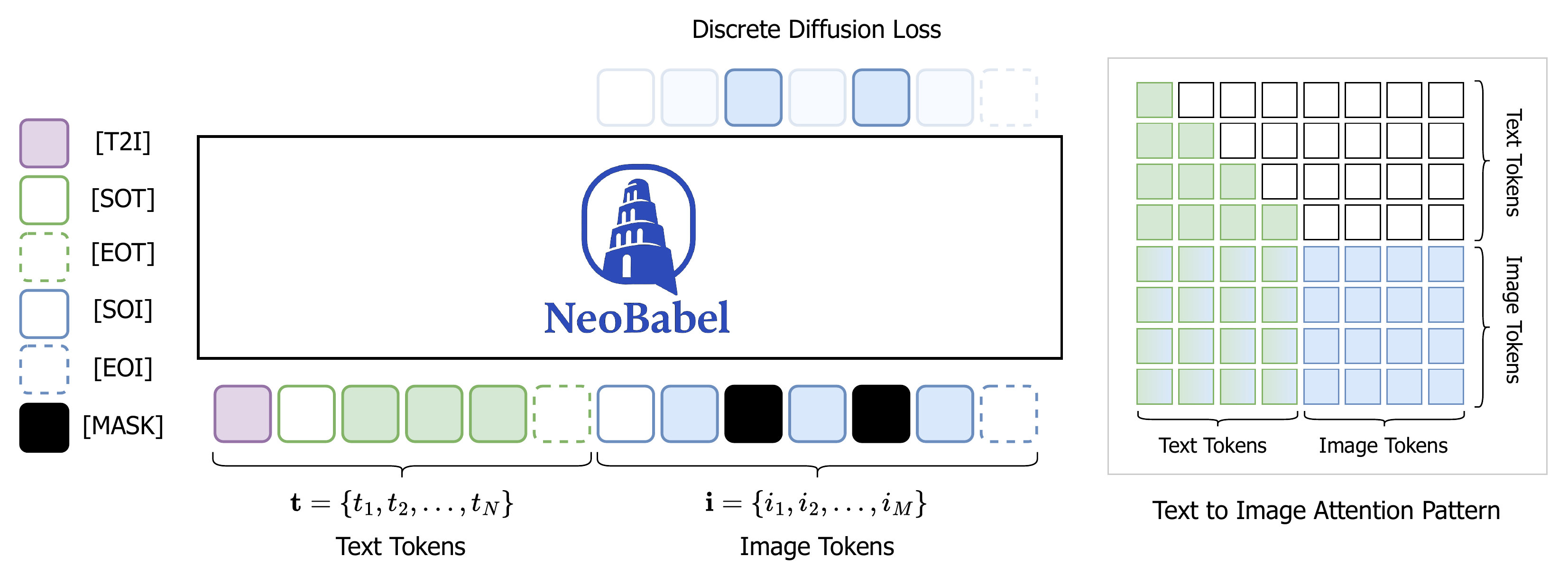}
    \caption{\textbf{NeoBabel: A Multilingual Open
Tower for Visual Generation.} Regardless of modality, all input data is first tokenized and embedded into a unified input sequence. \ours then applies causal attention to text tokens and full attention within a discrete denoising diffusion framework for image tokens, ultimately generating the desired image. This design enables \ours to support a wide range of tasks, including text-to-image generation, text-guided inpainting and extrapolation, as well as cross-lingual image generation.}
    \label{fig:neobabel_arch}
\end{figure}

Our architecture's core components, a multilingual tokenizer and transformer backbone, are specifically optimized for efficient, scalable cross-lingual image generation, supporting seamless processing across diverse languages and image types. Figure~\ref{fig:neobabel_arch} provides an overview of the \ours architecture. 

\subsubsection{Tokenizers}
\textbf{Text Tokenization.} For textual input, we adopt the tokenizer of the pretrained multilingual large language model Gemma-2 \citep{team2024gemma} without any modifications. This approach maintains compatibility with multilingual inputs while utilizing proven tokenization methods from language modeling.

\textbf{Image Tokenization.} For image input, we leverage the MAGVIT-v2 quantizer \citep{yu2023language} retrained by Show-o \citep{xie2024show} on 25 million images. This lookup-free quantizer learns a discrete codebook of size $K {=} 8{,}192$ and encodes $256 \times 256$ resolution images into $16 \times 16$ grids of discrete tokens. The quantization approach supports efficient downstream training and generation while preserving fine-grained visual details.

\subsubsection{Transformer Backbone}
As we build upon the pretrained multilingual large language model (LLM) Gemma-2 \citep{team2024gemma}, we maintain its overall transformer architecture, while introducing two key modifications: (1) integration of a unified multimodal embedding space, and (2) modality-aware attention patterns for flexible generation. Additionally, we apply qk-norm \citep{henry2020query} to each attention layer to enhance training stability and convergence.

\textbf{Unified Multimodal Embedding and Prompt Design.}
To enable seamless multimodal learning, we extend the LLM’s embedding table with 8,192 new learnable embeddings for discrete image tokens, allowing the model to process image inputs natively without architectural changes. Both text and image tokens are embedded in a shared space, enabling the model to learn cross-modal compositionality and semantic alignment. We represent all tasks including text-to-image generation as unified autoregressive sequences. Given a tokenized image-text pair, text and image tokens are concatenated into a single sequence. Special tokens such as \texttt{[T2I]}, \texttt{[SOT]}, \texttt{[EOT]}, \texttt{[SOI]}, and \texttt{[EOI]} explicitly mark task type and modality boundaries, enabling the model to disambiguate different modalities and tasks through prompting alone. This design simplifies the training pipeline by removing the need for modality-specific components or task-specific heads, allowing for flexible, scalable, and unified multimodal generation.

\textbf{Modality-Aware Attention Patterns.}
To accommodate the differing structural needs of text and image modalities, we employ a hybrid attention mechanism. Text tokens are modeled with causal attention to preserve autoregressive language modeling capabilities. Image tokens, in contrast, are modeled using full bidirectional attention, allowing rich interactions that are critical for high-fidelity image synthesis. When both modalities are present, attention masks are dynamically configured so that image tokens can fully attend to text tokens and preceding image tokens, enabling coherent, contextually grounded generation.

\subsection{Training Objective}
\label{sec:training-objective}
The model is trained on sequences composed of both textual and visual tokens, where text tokens act as a prefix and visual tokens form the postfix. We do not apply any learning objective to the text tokens; the loss is computed solely over the visual tokens.

Let $\mathbf{t} = \{t_1, t_2, \dots, t_N\}$ denote the text tokens and $\mathbf{i} = \{i_1, i_2, \dots, i_M\}$ denote the image tokens, forming a full input sequence $[\mathbf{t}; \mathbf{i}]$. During training, we randomly select a subset $\mathcal{J} \subset \{1, \dots, M\}$ of image token indices to be masked. The corresponding masked sequence is denoted by $\mathbf{i}_\ast$, where $i_j$ is replaced with a special \texttt{[MASK]} token for all $j \in \mathcal{J}$. The model is trained to reconstruct the original visual tokens at the masked positions by conditioning on the full input sequence of text tokens and (partially masked) image tokens. The objective is defined as:
\begin{equation}
\mathcal{L} = \sum_{j \in \mathcal{J}} \log p_\theta(i_j \mid \mathbf{t}, \mathbf{i}_\ast),
\end{equation}
where $p_\theta(\cdot)$ is the model's predicted distribution over image codebook entries, parameterized by $\theta$. The loss is only applied to the masked image tokens in $\mathcal{J}$. We follow the masking strategy introduced by \citet{xie2024show}, randomly masking a fixed ratio of visual tokens within each training sample. To further improve generation controllability, we incorporate classifier-free guidance~\citep{ho2022classifier} by replacing the conditioning text with a null string with some probability during training.

\subsection{Multilingual Model Merging}
\label{sec:model-merging-method}

To enhance generalization and stability of multilingual image generation models, we adopt model merging techniques that combine multiple checkpoints from the training trajectory. Let $\{M_i\}_{i=1}^{N}$ denote a sequence of $N$ model checkpoints and $\{w_i\}_{i=1}^{N}$ their corresponding non-negative weights. The merged model $\widehat{M}$ is defined as a convex combination:
\begin{equation}
\widehat{M} = \sum_{i=1}^{N} \alpha_i M_i \quad \text{where} \quad \alpha_i = \frac{w_i}{\sum_{j=1}^{N} w_j}.
\end{equation}
This formulation allows the merged model to interpolate within the solution space spanned by the selected checkpoints, potentially improving generalization on unseen prompts and enhancing robustness to overfitting. We consider three widely used weighting strategies for this purpose, each reflecting different assumptions about model evolution during training.
The comparative results and analysis of these approaches are presented ablation studies section.

\textbf{Simple Moving Average} (SMA) assigns equal weight to all checkpoints. It is defined as:
\begin{equation}
    M_{\text{avg}} = \frac{1}{N} \sum_{i=1}^{N} M_i.
\end{equation}
SMA is simple, stable, and particularly effective when applied in the later stages of training where model weights exhibit minimal drift. 
Prior work~\citep{li2025model} found SMA to perform robustly due to this stabilization.

\textbf{Exponential Moving Average} (EMA) emphasizes recent checkpoints by applying exponentially decaying weights. It is computed recursively as:
\begin{equation}
    M^{(i)}_{\text{avg}} = \alpha M_i + (1 - \alpha) M^{(i-1)}_{\text{avg}}, \quad i \in [2, N].
\end{equation}
The decay factor $\alpha \in (0,1)$ controls the trade-off between recency and stability. EMA adapts more quickly to recent model dynamics but is sensitive to noise if weights are unstable.

\textbf{Weighted Moving Average} (WMA) assigns custom, possibly increasing weights to later checkpoints. The merged model is computed using the normalized form:
\begin{equation}
    M_{\text{avg}} = \sum_{i=1}^{N} \frac{w_i}{\text{w}_{\text{sum}}} M_i, \quad \text{where} \quad \text{w}_{\text{sum}} = \sum_{i=1}^{N} w_i.
\end{equation}
This general formulation allows flexibility in how much importance is placed on each checkpoint. In our case, we use $w_i = i$ to emphasize later-stage models.

\section{NeoBabel Multilingual Datasets}
\label{sec:neobabel_multilingual_datasets}

\subsection{Data Curation Pipeline}
\label{sec:data_curation_pipeline}

\begin{table}[t]
    \centering
    \small
    \resizebox{\textwidth}{!}{%
    \begin{tabular}{lllrccr}
    \toprule
     \multicolumn{4}{c}{\textbf{Original English-Only Dataset}} &      \multicolumn{3}{c}{\textcolor{DarkBlue}{\textbf{\ours Multilingual Expansion}}} \\
    \cmidrule(lr){1-4}  \cmidrule(lr){5-7}
    Dataset  & Image Source & Caption Source & Size &  \textcolor{DarkBlue}{Recaptioning} & \textcolor{DarkBlue}{Translation} & \textcolor{DarkBlue}{New Size} \\
    \midrule
    ImageNet 1K & Web & Class labels &  1M & \textcolor{DarkBlue}{--} & \textcolor{DarkBlue}{\checkmark} & \textcolor{DarkBlue}{6M}\\
    %\arrayrulecolor{gray}\midrule\arrayrulecolor{black}
    CC12M  &  Web  & Alt-text (noisy) & 12M & \makecell{\textcolor{DarkBlue}{\checkmark}} & \textcolor{DarkBlue}{--}& \textcolor{DarkBlue}{12M}\\
    %\arrayrulecolor{gray}\midrule\arrayrulecolor{black}
    %\arrayrulecolor{gray}\midrule\arrayrulecolor{black}
    SA-1B  & Photography & LLaVA  & 10M &  \makecell{\textcolor{DarkBlue}{\checkmark}} & \textcolor{DarkBlue}{--}& \textcolor{DarkBlue}{10M} \\
    
    LAION-Aesthetic   & Web & Alt-text (noisy)  & 12M & \makecell{\textcolor{DarkBlue}
    {\checkmark}} & \textcolor{DarkBlue}{\checkmark}& \textcolor{DarkBlue}{72M} \\    %\arrayrulecolor{gray}\midrule\arrayrulecolor{black}
    JourneyDB  & Synthetic & GPT-3.5  & 4M & \makecell{\textcolor{DarkBlue}{\checkmark}} & \textcolor{DarkBlue}{\checkmark} &  \textcolor{DarkBlue}{24M} \\
    %\arrayrulecolor{gray}\midrule\arrayrulecolor{black}
    BLIP3-o Instruct & Web + Synthetic & GPT-4o / human   & 60K & \textcolor{DarkBlue}{--} & \textcolor{DarkBlue}{\checkmark} & \textcolor{DarkBlue}{360K} \\
    \midrule
    & & & 39M & & & \textcolor{DarkBlue}{124M}\\
    \bottomrule
    \end{tabular}%
    }
    \caption{\textbf{NeoBabel multilingual datasets}, detailing their English-only data source, image origin, caption format, and size. Our multilingual expansion covers model-generated recaptioning, translation into multiple languages, or both. Our expansions increase the total size from 39M to 124M image–caption/label pairs. In the remainder of this paper, all modified datasets are prefixed with m- to denote their expanded form.}
    \label{tab:data_overview}
\end{table}

Multilingual multimodal data remains scarce, especially compared to the abundance of English-centric resources. This imbalance poses a significant barrier to training and evaluating models that can understand grounded language across diverse linguistic contexts. 
To address this gap, we curate and augment several multilingual datasets by translating and recaptioning existing image-caption pairs into six target languages: \texttt{English}, \texttt{Chinese}\footnote{Throughout this work `Chinese' refers to Simplified Chinese.}, \texttt{Dutch}, \texttt{French}, \texttt{Hindi}, and \texttt{Persian}.
We summarize the datasets curated in Table \ref{tab:data_overview}.
At the core of our approach is a multilingual captioning pipeline designed to ensure both semantic richness and linguistic diversity. 
We begin by generating a detailed English caption for each image using InternVL \citep{chen2024internvl}, prompted with a simple instruction: ``Describe this image in detail in English.'' This step guarantees comprehensive coverage of the visual content. 

To preserve quality and consistency across languages, we implement a multi-step post-processing and filtering stage based on four strategies:
\begin{itemize}
    \item \textbf{Length filtering}: Remove captions that are too short (e.g., fewer than 5 tokens) or excessively long (e.g., more than 500 tokens).
    \item \textbf{Language validation}: Detect and discard captions containing non-English phrases or corrupted outputs using language identification tools. We use the \texttt{fastText} language identification model trained on 176 languages \citep{joulin2016fasttext}. We discard any caption not classified as English with a confidence score above 90\%.
    \item \textbf{Visual-text mismatch filtering}: Discard captions that do not align with visual content, measured via auxiliary vision-language models (e.g., using VQAScore). Specifically, we leverage MolMo-72B \citep{deitke2024molmo} deployed with vLLM \citep{kwon2023efficient}, formulating the task as a binary structured prediction (\texttt{yes}/\texttt{no}) via vLLM's output interface.
    \item \textbf{Toxicity and NSFW filtering}: Discard samples using the LAION-5B NSFW classifier \citep{schuhmann2022laion} to ensure safe visual content before captioning, assuming high likelihood of appropriateness in the resulting captions.
\end{itemize}
Once high-quality English captions are obtained, we translate them into five target languages using the NLLB model~\citep{costa2022no} for the pretraining datasets, and the Gemini Experimental model (gemini-2.0-flash-lite) for the instruction tuning datasets. This separation ensures high translation coverage at scale during pretraining, while leveraging higher-quality outputs for instruction-tuned data. Using English as a pivot allows us to take advantage of strong captioning performance in high-resource settings while ensuring consistent semantic content across all languages. 
This approach not only amplifies the linguistic diversity of our dataset but also maintains alignment between captions, which is critical for multilingual training and evaluation. Ultimately, this step plays a central role in constructing a high-quality, language-balanced multimodal resource---an essential step toward more inclusive and globally-relevant vision-language models.

\subsection{\ours Pretraining Data}

The previous section described the overall pipeline and transformation steps, and next we detail the data sources and multilingual adaptations used to train the model.
We use a diverse collection of image-text datasets to build strong multilingual visual-language alignment combining real-world and synthetic image sources.
While the images are drawn from established, high-quality datasets, the accompanying captions have been significantly enriched through our recaptioning and multilingual translation pipeline—resulting in a more diverse, detailed, and valuable resource for future multilingual generative models.

\textbf{m-ImageNet-1K:} The original English class labels are translated into five more languages to obtian a total of six target languages, forming multilingual textual prompts for class-conditional image generation.

\textbf{m-SA-1B and m-CC12M:} We incorporate 22 million image-caption pairs in English from SA-1B \citep{kirillov2023segment} and CC12M \citep{changpinyo2021conceptual}. These datasets provide rich natural image-caption pairs and enhance visual diversity. The texts are enhanced through our recaptioning pipeline described in Section \ref{sec:data_curation_pipeline}.

\textbf{m-LAION-Aesthetic:} A subset of the LAION dataset including 12M image-text pairs\footnote{\url{https://huggingface.co/datasets/dclure/laion-aesthetics-12m-umap}} is enhanced and translated, yielding approximately 72 million image-caption pairs for a total of six languages. 

\textbf{m-JourneyDB:} This synthetic dataset consists of 4 million high-quality images generated by the Midjourney model \citep{sun2023journeydb}. We apply the same recaptioning and translation pipeline to generate 24 million image-caption pairs for our six languages.

Combining all sources, the final pretraining dataset contains approximately 124 million image-text pairs across six languages, covering diverse domains and visual aesthetics.

\subsection{\ours Instruction Tuning Data}

Here we describe our datasets and mixing strategies used for instruction tuning. This phase reuses two datasets introduced earlier and adds a smaller but higher-quality dataset focused on multimodal instruction tuning:

\textbf{m-LAION-Aesthetic and m-JourneyDB:} Our setup continues to use the LAION-Aesthetic and JourneyDB datasets, as extended in the pretraining data.

\textbf{m-BLIP3o-Instruct:} An instruction-focused dataset introduced by \citet{chen2025blip3}, containing multimodal instruction samples, also translated into six languages for multilingual supervision.

All images are resized to $512 \texttimes 512$. While the images are drawn from established, high-quality sources, most accompanying texts have been significantly enriched or rewritten, resulting in a more valuable and linguistically diverse dataset for instruction tuning and multilingual generation.

\section{NeoBabel Training Stages: Learning Progression}
\label{sec:neobabel_training_stages}

% In the following section, we outline how they are integrated into a staged learning framework, first for progressive pre-training and subsequently for multilingual instruction tuning, designed to maximize both visual fidelity and linguistic generalization.

\ours is trained using a staged learning framework consisting of three progressive pretraining stages (Section \ref{sec:progressive_pretraining})  followed by two instruction tuning stages (Section \ref{sec:progressive_instruction_tuning}).

\subsection{Progressive Pretraining}
\label{sec:progressive_pretraining}

Our pretraining includes three stages, progressively scaling from basic visual understanding to advanced multilingual image generation:

\textbf{Stage 1 -- Pixel Dependency Learning:}
The model initially learns foundational visual representations using m-ImageNet-1K. Class-conditional image generation is guided by translated class labels, enabling the model to form robust image token embeddings and capture pixel-level dependencies for high-fidelity output.

\textbf{Stage 2 -- Scaling Alignment with Large-Scale Multilingual Data:}
Using weights from the first stage, the model is fine-tuned on 22 million English-only image-caption pairs (from m-SA-1B and m-CC12M) and 72 million translated samples from m-LAION-Aesthetic. This stage strengthens the model’s grounding in natural image-text alignment while developing multilingual capabilities through broad cross-lingual exposure.

\textbf{Stage 3 -- Refined Multilingual Pretraining:}
In the final stage, the model is trained on 96 million multilingual image-text pairs derived from m-LAION-Aesthetic and m-JourneyDB. The training balances high-quality real-world aesthetic data with diverse, synthetic images to improve generalization across languages, domains, and modalities.

\subsection{Progressive Instruction Tuning}
\label{sec:progressive_instruction_tuning}

Following pretraining, the model advances to instruction tuning, where the focus shifts from unsupervised representation learning to explicit task-guided adaptation, refining its ability to interpret and execute complex, multilingual instructions through our curated datasets and progressive exposure to prompt-driven generation in two stages:

\textbf{Stage 1 -- Initial Multilingual Instruction Alignment:}
To build robust multilingual instruction-following capabilities at high resolution, the model is first trained with a diverse mixture of the three datasets described above. In this stage, training samples are drawn from m-LAION-Aesthetic, m-JourneyDB, and m-BLIP3o-Instruct using mixing weights $\alpha_1$, $\alpha_2$, and $\alpha_3$, respectively, such that $\alpha_1 + \alpha_2 + \alpha_3 = 100$. A higher $\alpha_1$ and moderate $\alpha_2$ prioritize real-world and aesthetic content, while a smaller $\alpha_3$ introduces early exposure to instruction-rich samples. This balance helps the model learn cross-lingual, cross-modal grounding without overwhelming it with complex prompts in the early stages.

\textbf{Stage 2 -- Instruction Refinement:}
In the second stage, we adjust the mixing weights to emphasize instruction-rich and synthetic supervision. Specifically, $\alpha_2$ and $\alpha_3$ are increased to draw more heavily from m-JourneyDB and m-BLIP3o-Instruct, while $\alpha_1$ is decreased to reduce reliance on LAION-based content. This curriculum-style shift enables the model to refine its instruction-following capabilities using complex multilingual prompts and high-quality synthetic images. The increased semantic richness improves the model's generalization to both benchmark instruction tasks and open-ended generation scenarios.

Each stage is trained for 500k steps (except the final stage of instruction tuning with 200k) using the AdamW optimizer and cosine learning rate decay. The learning rate is set to $1\text{e}{-4}$ during pretraining and adjusted during instruction tuning. We gradually increase prompt sequence length and resolution from $128$ to $512$ and from $256 \times 256$ to $512\times512$ respectively. The vocabulary and codebook sizes are fixed across all stages. Full hyperparameter settings for each pretraining and instruction tuning stage are summarized in the Appendix. %Table~\ref{tab:hyperparameters}.

\section{Multilingual Evaluation of Image Generation}
\label{sec:eval_suite}

Existing image generation benchmarks are mostly English-centric, failing to capture cross-lingual performance. To resolve this limitation, we introduce a multilingual evaluation suite that extends established (English-only) benchmarks to cover six diverse languages and introduces new evaluation metrics for assessing cross-lingual visual consistency. This section outlines our multilingual evaluation suite (Section~\ref{sec:multilingual_eval_sets}) and multilingual evaluation metrics (Section~\ref{sec:multilingual_eval_metrics}).

\subsection{Multilingual Evaluation Suite}
\label{sec:multilingual_eval_sets}

We assess the image generation capabilities of \ours using two complementary benchmarks: {GenEval}~\citep{ghosh2023geneval} and {DPG-Bench}~\citep{hu2024ella}. GenEval offers a structured evaluation of prompt-to-image alignment across six compositional dimensions: \textit{single object}, \textit{two objects}, \textit{counting}, \textit{colors}, \textit{position}, and \textit{color attribute}. In contrast, DPG-Bench targets general-purpose generation with open-ended, diverse prompts that test broader semantic understanding. However, both benchmarks are English-only and fail to capture multilingual generative performance.

As part of our multilingual evaluation suite, we introduce \textbf{m-GenEval} and \textbf{m-DPG}, multilingual extensions of the original benchmarks. All prompts are translated into five additional languages: \texttt{Chinese}, \texttt{Dutch}, \texttt{French}, \texttt{Hindi}, and \texttt{Persian}, using the Gemini Experimental model, followed by human verification and manual corrections to ensure semantic fidelity and linguistic fluency. Together with the paper, we  publicly release m-GenEval and m-DPG to promote inclusive and realistic evaluation of multilingual text-to-image models and support broader community adoption.

\subsection{Multilingual Evaluation Metrics}
\label{sec:multilingual_eval_metrics}

To complement the multilingual benchmarks introduced above, we introduce two metrics that assess how well generative models preserve visual and semantic consistency across languages. Existing evaluations focus on monolingual alignment, overlooking whether models produce consistent outputs across languages or under mixed-language inputs. To address this, we introduce two scores to assess cross-lingual consistency and robustness under intra-prompt language mixing. Together, these metrics provide a more diagnostic view of multilingual generation performance.

\textbf{Cross-Linguistic Consistency (CLC).} To evaluate whether multilingual models generate semantically consistent and faithful outputs across languages, we introduce the CLC score. 
Multilingual image generation models should produce visually similar outputs when given semantically equivalent prompts, regardless of the input language. Measuring this consistency is crucial for understanding how well the model aligns its multilingual text inputs with the corresponding visual outputs, which reflects the quality of its cross-lingual grounding.
We evaluate in a multilingual setting consisting of $P$ prompts, each paired with $L$ language variations, forming a parallel dataset.
For each prompt $p \in \{p_i\}_{i=1}^{P}$, we generate $K$ images (one per language), resulting in $L \times K$ images. Let $x_i$ denote an image and $f(x_i) \in \mathbb{R}^d$ its corresponding embedding obtained from a vision encoder. 

To measure consistency, we treat the $K$ images generated from the English version of the prompt as the reference set $\mathcal{R}_p$, and the remaining $(L - 1) \times K$ images generated from other languages as the target set $\mathcal{T}_p$. The core idea is that if the model is truly language-agnostic in its understanding, images generated from non-English prompts should be visually similar to those generated from the English prompt.
The CLC Score for prompt $p$ is computed by averaging the cosine similarity between all reference and non-reference embeddings:
\begin{equation}
\mathrm{CLC}_p = \frac{1}{|\mathcal{R}_p| \cdot |\mathcal{T}_p|} \sum_{x_i \in \mathcal{R}_p} \sum_{x_j \in \mathcal{T}_p} \cos\left(f(x_i), f(x_j)\right).
\end{equation}
Finally, the overall CLC score is obtained by averaging $\mathrm{CLC}_p$ over all prompts $P$. For evaluation, we use m-DPG prompts and compute embeddings with two strong vision encoders, EVA-CLIP \citep{sun2023eva} and DINOv2 \citep{oquab2023dinov2}, to ensure robustness across different feature representations.
This metric provides a quantitative measure of how well multilingual generation models maintain semantic and visual alignment across languages. %, an important step toward truly universal and language-agnostic multimodal AI systems.

\textbf{Code-Switching Similarity (CSS).} 
Real-world multilingual communication frequently involves code switching, i.e., interleaving of multiple languages within a single utterance. 
Therefore, a well-aligned multilingual model should demonstrate robustness not only to monolingual prompts but also to mixed-language inputs, capturing the inherent complexity and variability of natural language.
Code switching often increases perplexity and degrades performance in language models; however, its impact on image generation remains largely unexplored.
%While code switching often increases perplexity and degrades performance in language models, its impact on image generation remains underexplored. 
To evaluate this, we introduce the CSS Score, which quantifies visual consistency under intra-prompt language variation. 
Given a set of reference prompts composed entirely in English, we construct two variants per prompt for each of the $L{-}1$ non-English target languages: (1) \texttt{English-First (EF)}: the first half of the prompt remains in English while the second half is translated into the target language, and (2) \texttt{English-Second (ES)}: the first half is translated while the second half remains in English.

For each  prompt $p \in \{p_i\}_{i=1}^{P}$, we generate a single reference image $x_\text{ref}$ from the original English prompt and $L-1$ code-switched images: $x_\text{EF}^{(l)}$ and $x_\text{ES}^{(l)}$ for each target language $l$. Each image is encoded into an embedding $f(x) \in \mathbb{R}^d$ using a vision encoder. The Code Switching Similarity (CSS) score for each prompt is computed by measuring the average cosine similarity between the reference embedding $f(x_\text{ref})$ and the embeddings from the EF and ES variants:
\begin{equation}
\mathrm{CSS}{p}^{\text{EF}} = \frac{1}{L-1} \sum_{l=1}^{L-1} \cos\left(f(x_\text{ref}), f(x_\text{EF}^{(l)})\right), \quad
\mathrm{CSS}{p}^{\text{ES}} = \frac{1}{L-1} \sum_{l=1}^{L-1} \cos\left(f(x_\text{ref}), f(x_\text{ES}^{(l)})\right).
\end{equation}
The final CSS scores are obtained by averaging across all prompts:
\begin{equation}
\mathrm{CSS}^{\text{EF}} = \frac{1}{P} \sum_{p=1}^{P} \mathrm{CSS}_{p}^{\text{EF}}, \quad
\mathrm{CSS}^{\text{ES}} = \frac{1}{P} \sum_{p=1}^{P} \mathrm{CSS}_{p}^{\text{ES}}.
\end{equation}
To assess how well models preserve semantic consistency under intra-prompt code switching, we report both $\mathrm{CSS}^{\text{EF}}$ and $\mathrm{CSS}^{\text{ES}}$, using embeddings from EVA-CLIP \citep{sun2023eva} and DINOv2 \citep{oquab2023dinov2} computed on m-DPG prompts. 
\begin{table}[t]
\centering
\footnotesize
\renewcommand{\arraystretch}{1.1}
\setlength{\tabcolsep}{4pt}
\resizebox{\textwidth}{!}{%
\begin{tabular}{lccrcccccc|c}
\toprule
\textbf{Method} & \faGlobe & \textbf{Type} & \textbf{Params.} &
\makecell[c]{\textbf{Single} \\ \textbf{Object}} &
\makecell[c]{\textbf{Two} \\ \textbf{Object}} &
\textbf{Counting} &
\textbf{Colors} &
\textbf{Position} &
\makecell[c]{\textbf{Color} \\ \textbf{Attribute}} &
\textbf{Overall} \\
\midrule
LlamaGen      & \texttimes & G     & 0.8B  & 0.71 & 0.34 & 0.21 & 0.58 & 0.07 & 0.04 & 0.32 \\
LDM       & \texttimes & G     & 1.4B  & 0.92 & 0.29 & 0.23 & 0.70 & 0.02 & 0.05 & 0.37 \\
SDv1.5    & \texttimes & G     & 0.9B  & 0.97 & 0.38 & 0.35 & 0.76 & 0.04 & 0.06 & 0.43 \\
PixArt-alpha & \texttimes & G     & 0.6B  & 0.98 & 0.50 & 0.44 & 0.80 & 0.08 & 0.07 & 0.48 \\
SDv2.1    & \texttimes & G     & 0.9B  & 0.98 & 0.51 & 0.44 & 0.85 & 0.07 & 0.17 & 0.50 \\
DALL-E 2  & \texttimes & G     & 6.5B  & 0.98 & 0.66 & 0.49 & 0.77 & 0.10 & 0.19 & 0.52 \\
SDXL       & \texttimes & G     & 2.6B  & 0.98 & 0.74 & 0.39 & 0.85 & 0.15 & 0.23 & 0.55 \\
SD3         & \texttimes & G     & 2B    & 0.98 & 0.74 & 0.63 & 0.67 & 0.34 & 0.36 & 0.62 \\
\midrule
CoDI         & \texttimes & U\&G & -  & 0.89 & 0.16 & 0.16 & 0.65 & 0.02 & 0.01 & 0.31 \\
Chameleon       & \texttimes & U\&G & 7B    & -    & -    & -    & -    & -    & -    & 0.39 \\
LWM          & $\circ$ & U\&G & 7B    & 0.93 & 0.41 & 0.46 & 0.79 & 0.09 & 0.15 & 0.47 \\
SEED-X         & $\circ$ & U\&G & 17B   & 0.97 & 0.58 & 0.26 & 0.80 & 0.19 & 0.14 & 0.49 \\
Janus          & $\circ$ & U\&G & 1.3B  & -    & -    & -    & -    & -    & -    & 0.61 \\
TokenFlow          & $\circ$ & U\&G & 14B  & -    & -    & -    & -    & -    & -    & 0.63 \\
EMU3          & $\circ$ & U\&G & 8B  & -    & -    & -    & -    & -    & -    & 0.66 \\
Show-o          & \texttimes & U\&G & 1.3B  & 0.98 & 0.80 & \textbf{0.66} & 0.84 & 0.31 & 0.50 & 0.68 \\
Janus-Pro          & $\circ$ & U\&G & 7B  & -    & -    & -    & -    & -    & -    & 0.80 \\
BLIP3-o         & $\circ$ & U\&G   & 4B  & - & - & - & - & - & - & 0.81 \\
BLIP3-o         & $\circ$ & U\&G   & 8B  & - & - & - & - & - & - & \textbf{0.83} \\
\midrule
% \midrule
\rowcolor{LightBlue}
\ours           & \checkmark & G & 2B & \textbf{1.00}	& \textbf{0.91} & 0.62 &	\textbf{0.91} &	\textbf{0.81} &	\textbf{0.77} & \textbf{0.83} \\
\bottomrule
\end{tabular}
}
\caption{\textbf{English-only GenEval benchmark comparison.} \ours achieves the highest overall score, outperforming larger models on tasks requiring compositional reasoning and fine-grained prompt-image alignment. Symbol legend: \faGlobe~denotes multilingual generation capability, with \checkmark~indicates a full multilingual capability, $\circ$~represents partial multilingual capability (i.e. bilingual or multilingual to a limited extent), and \texttimes~denotes monolingual models. 
}
\label{tab:geneval_enlgish}
\end{table}

\section{Results and Discussions}
\label{sec:experiment}

We evaluate \ours on our multilingual extension of standard benchmarks, including m-GenEval and m-DPG, to assess performance across languages both quantitatively and qualitatively. 

\textbf{Baselines.} We evaluate our model against a diverse range of baselines, which we group into two categories: generative-only models (G) and unified models (U\&G). The generative models are designed exclusively for text-to-image generation, without any visual understanding components. This category includes LlamaGen \citep{sun2024autoregressive}, LDM \citep{rombach2022high}, SDv1.5 and SDv2.1 \citep{rombach2022high}, SDXL \citep{Podell2023SDXLIL}, SD3 \citep{esser2024scaling}, DALL-E 2 \citep{ramesh2022hierarchical}, and PixArt-$\alpha$ \citep{pixart} models primarily optimized for high-quality and compositional image generation. In contrast, the unified models support both image generation and image understanding tasks such as captioning and visual question answering. This group includes CoDI \citep{tang2023any}, LWM \citep{liu2024world}, SEED-X \citep{ge2024seed}, Chameleon \citep{team2024chameleon}, TokenFlow \citep{qu2025tokenflow}, EMU3 \citep{wang2024emu3}, Janus \citep{wu2025janus}, Janus-Pro \citep{chen2025janus}, and BLIP3-o \citep{chen2025blip3}. Our comparison includes both small-scale and large-scale models, spanning from under 1B to over 17B parameters.

\begin{figure}[t!]
    \centering
    \includegraphics[width=1\linewidth, trim=0 45 0 0, clip]{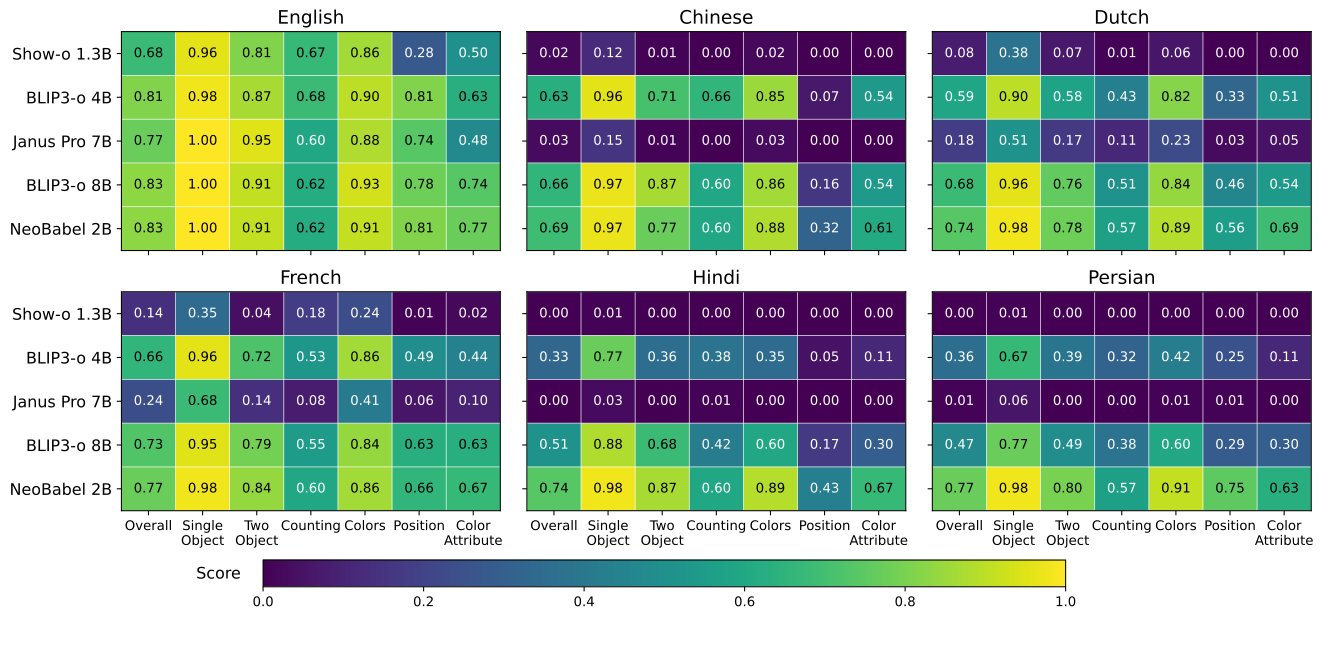}
    \caption{\textbf{m-GenEval benchmark comparison.} Models such as Janus Pro and BLIP3-o rely on multilingual base LLMs but are trained solely on English image-generation data, leading to a sharp performance drop in non-English languages. In contrast, \ours maintains strong and consistent results across all six languages, demonstrating robust cross-lingual generalization. Here baseline models are ordered by parameter count. 
    }
    \label{fig:multilingual_geneval}
\end{figure}

\subsection{Multilingual Image Generation Performance}
\label{sec:qual_quant}
\textbf{m-GenEval Comparison.} We begin by evaluating \ours on the English prompts of the m-GenEval benchmark, with results reported in Table~\ref{tab:geneval_enlgish}. The comparison includes both generative models (G), which focus solely on text-to-image generation, and unified models (U\&G), which also support image understanding tasks such as captioning and visual question answering. Despite having only 2B parameters, \ours outperforms or matches best-performing unified models such as Janus-Pro 7B (0.77) and BLIP3-o 8B (0.83), which are significantly larger in terms of parameters. It also surpasses SD3 2B (0.62), a leading model in the generative category, achieving the highest overall score of 0.83. This performance reflects strong fine-grained and compositional prompt-image alignment particularly in challenging subcategories like color attributes and positional grounding.

In Figure~\ref{fig:multilingual_geneval}, we further evaluate \ours across five more languages including \texttt{Chinese}, \texttt{Dutch}, \texttt{French}, \texttt{Hindi}, and \texttt{Persian} to assess its multilingual generalization capabilities beyond English. As can be seen, the performance gap between \ours and the strongest baselines is small in Chinese (by 0.03). We attribute this to two factors: (i) the use of bilingual English-Chinese instruction-tuning data in models like Janus Pro, whose training setup is not publicly disclosed, and (ii) architectural choices such as BLIP3-o's use of a frozen LLM backbone with prompt learning instead of full model adaptation. In medium-resource languages like Dutch and French, the gap widens (0.06 and 0.04 respectively), and in low-resource languages such as Hindi and Persian, \ours significantly outperforms all baselines by a large margin (up to 0.3 improvement), despite having 4× fewer parameters than Janus Pro 7B and BLIP3-o 8B. These results underscore the cross-lingual robustness and data efficiency of our multilingual instruction-tuning strategy.

%Baseline models trained only on English-centric corpus, such as Janus Pro, BLIP3-o, and Show-o, show significant performance drops in other languages, with overall scores often falling below 0.3. In contrast, \ours maintains consistent performance across all languages, with scores ranging from 0.69 in Chinese to 0.77 in French and Persian. This stability highlights the robust instruction following ability of \ours and its reliable alignment across diverse linguistic inputs, establishing a strong baseline on m-GenEval.

\textbf{m-DPG Comparison.}
Compared to m-GenEval, which emphasizes fine-grained attributes and atomic compositional reasoning, m-DPG focuses on a model’s ability to follow natural, descriptive multilingual prompts. It tests whether the generated images are semantically accurate, detailed, and coherent, making it a stronger indicator of real-world prompt-image alignment performance.
We evaluate \ours on m-DPG to assess prompt-image alignment across 6 languages in Table~\ref{tab:dpg-bench}. \ours achieves comparable performance in English (0.75), even though it uses only 2B parameters, which is far fewer than BLIP3-o (4B and 8B) and Janus Pro (7B). More importantly, \ours outperforms all baselines in the non-English settings. Existing models show notable performance drops in several languages, especially in low-resource settings (Hindi and Persian) where models such as Janus, Janus Pro, and Show-o perform poorly. In contrast to the best-performing baseline (BLIP3-o 8B), \ours consistently achieves the highest scores across all six target languages. As in m-GenEval, we observe a similar trend in m-DPG, where the performance gap widens in medium-resource languages by 0.10 in Dutch and 0.09 in French and becomes even larger in low-resource settings, with gaps of 0.13 in Hindi and 0.12 in Persian.

% \begin{table}[t]
% \centering
% \caption{\textbf{Multilingual performance on the DPG-Bench benchmark.} \ours achieves competitive results in English and consistently outperforms all baselines across six non-English languages, demonstrating strong cross-lingual prompt understanding and image generation.}
% \begin{tabular}{lccccccc}
% \toprule
% \textbf{Model} & \textbf{Parameters} & \textbf{English} & \textbf{Dutch} & \textbf{French} & \textbf{Chinese} & \textbf{Hindi} & \textbf{Persian} \\
% \midrule
% Show-o & 1.3B     & 67.3 & 22.1	& 32.6	& 10.1	& 4.5 & 4.9\\
% EMU3 &  8B         & 80.6 & -- & -- & -- & -- & --   \\
% TokenFlow-XL & 14B& 73.4 & -- & -- & -- & -- & --   \\
% Janus & 1.3B      & 79.7 & 42.3	& 53.8 & 	56.4 &	17.8	& 13.8\\
% Janus Pro & 7B    & \textbf{84.2} & 61.7 &	68.8 &	50.0	& 12.4 & 12.4\\
% BLIP3-o & 4B      & 79.4 & 58.9 & 59.7 & 60.8 & 47.4 & 49.5 \\
% BLIP3-o & 8B      & 80.7 & 59.9	& 61.3	& 56.4	& 50.4	& 53.6\\
% \midrule
% \rowcolor{green!10}
% \ours &  2B        & 75.6 & \textbf{69.7}	& \textbf{70.7}	& \textbf{70.2}	& \textbf{63.9}	& \textbf{65.6}\\
% \bottomrule
% \end{tabular}
% \label{tab:dpg-bench}
% \end{table}

\begin{table}[t]
\centering
\resizebox{\textwidth}{!}{%
\begin{tabular}{lrcccccc|c}
\toprule
\textbf{Model} & \textbf{Params.} & \textbf{English} & \textbf{Chinese} & \textbf{Dutch} & \textbf{French} & \textbf{Hindi} & \textbf{Persian} & \textbf{Overall} \\
\midrule
Show-o          & 1.3B    & 0.67 & 0.10 & 0.22	& 0.32	& 0.04 & 0.04 & 0.23 \\
EMU3            &  8B     & 0.80 & -- & -- & -- & -- & --   & - \\
TokenFlow-XL    & 14B     & 0.73 & -- & -- & -- & -- & --   & - \\
Janus           & 1.3B    & 0.79 & 	0.56  & 0.42	& 0.53 &	0.17	& 0.13 & 0.43 \\
Janus Pro       & 7B      & \textbf{0.84} &	0.50	& 0.61 &	0.68 & 0.12 & 0.12 & 0.47 \\
BLIP3-o         & 4B      & 0.79 & 0.60 & 0.58 & 0.59 & 0.47 & 0.49 & 0.58  \\
BLIP3-o         & 8B      & 0.80 & 0.56	& 0.59	& 0.61 & 0.50	& 0.53 & 0.59 \\
\midrule
\rowcolor{LightBlue}
\ours &  2B        & 0.75 & \textbf{0.70}	& \textbf{0.69}	& \textbf{0.70}	& \textbf{0.63}	& \textbf{0.65} & \textbf{0.68} \\
\bottomrule
\end{tabular}
}
\caption{\textbf{m-DPG benchmark comparison.} Despite its small parameter count, \ours achieves competitive results in English and consistently outperforms all baselines across five non-English languages, demonstrating strong cross-lingual prompt understanding and image generation.}
\label{tab:dpg-bench}
\end{table}

% \begin{table}[t]
% \centering
% \begin{tabular}{lrcccccc|c}
% \toprule
% \textbf{Model} & \textbf{Params.} & \textbf{English} & \textbf{Chinese} & \textbf{Dutch} & \textbf{French} & \textbf{Hindi} & \textbf{Persian} & \textbf{Overall} \\
% \midrule
% Show-o          & 1.3B    & 0.67 & 0.10 & 0.22	& 0.32	& 0.04 & 0.04 & 0.23 \\
% EMU3            &  8B     & 0.80 & -- & -- & -- & -- & --   & - \\
% TokenFlow-XL    & 14B     & 0.73 & -- & -- & -- & -- & --   & - \\
% Janus           & 1.3B    & 0.79 & 	0.56  & 0.42	& 0.53 &	0.17	& 0.13 & 0.43 \\
% Janus Pro       & 7B      & \textbf{0.84} &	0.50	& 0.61 &	0.68 & 0.12 & 0.12 & 0.47 \\
% BLIP3-o         & 4B      & 0.79 & 0.60 & 0.58 & 0.59 & 0.47 & 0.49 & 0.58  \\
% BLIP3-o         & 8B      & 0.80 & 0.56	& 0.59	& 0.61 & 0.50	& 0.53 & 0.59 \\
% \midrule
% \rowcolor{LightBlue}
% \ours &  2B        & 0.75 & \textbf{0.70}	& \textbf{0.69}	& \textbf{0.70}	& \textbf{0.63}	& \textbf{0.65} & \textbf{0.68} \\
% \bottomrule
% \end{tabular}
% \caption{\textbf{Multilingual DPG-Bench comparision.} Despite its small parameter count, \ours achieves competitive results in English and consistently outperforms all baselines across five non-English languages, demonstrating strong cross-lingual prompt understanding and image generation.}
% \label{tab:dpg-bench}
% \end{table}

%\subsection{Qualitative Results}
\begin{figure}[th!]
    \centering
    \includegraphics[width=0.95\linewidth]{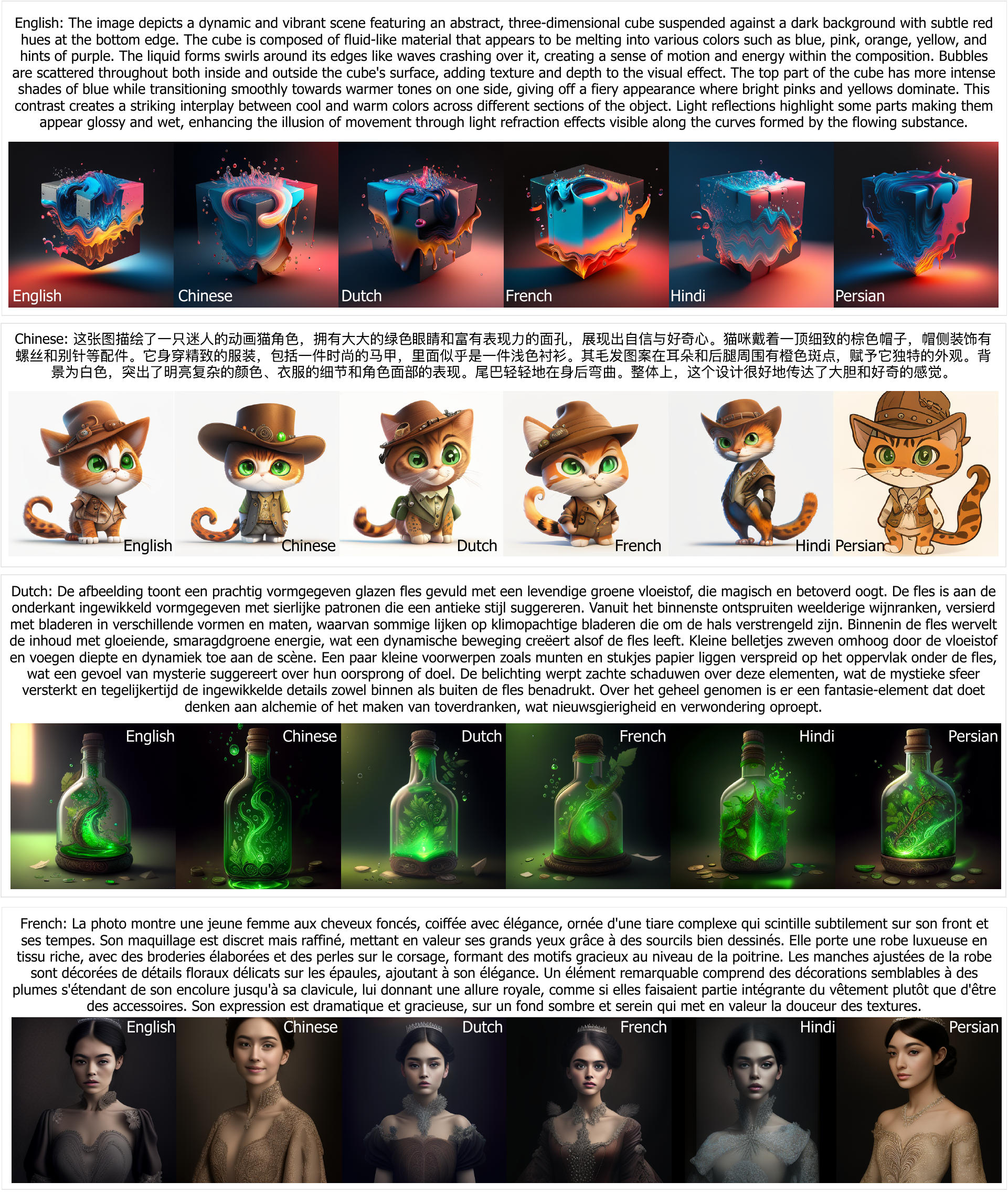}
    \caption{\textbf{Qualitative evaluation of \ours.} Each row is based on a single concept expressed in six different languages. For clarity, we show only one of the prompts (in one language) and present six images generated from its translated prompts in the other five languages. Across all languages, \ours delivers semantically accurate and visually cohesive outputs with reliable consistency.}
    \label{fig:qual-eval-1}
\end{figure}

\subsection{Qualitative Evaluation} 

\begin{figure}[th!]
    \centering
    \includegraphics[width=0.95\linewidth]{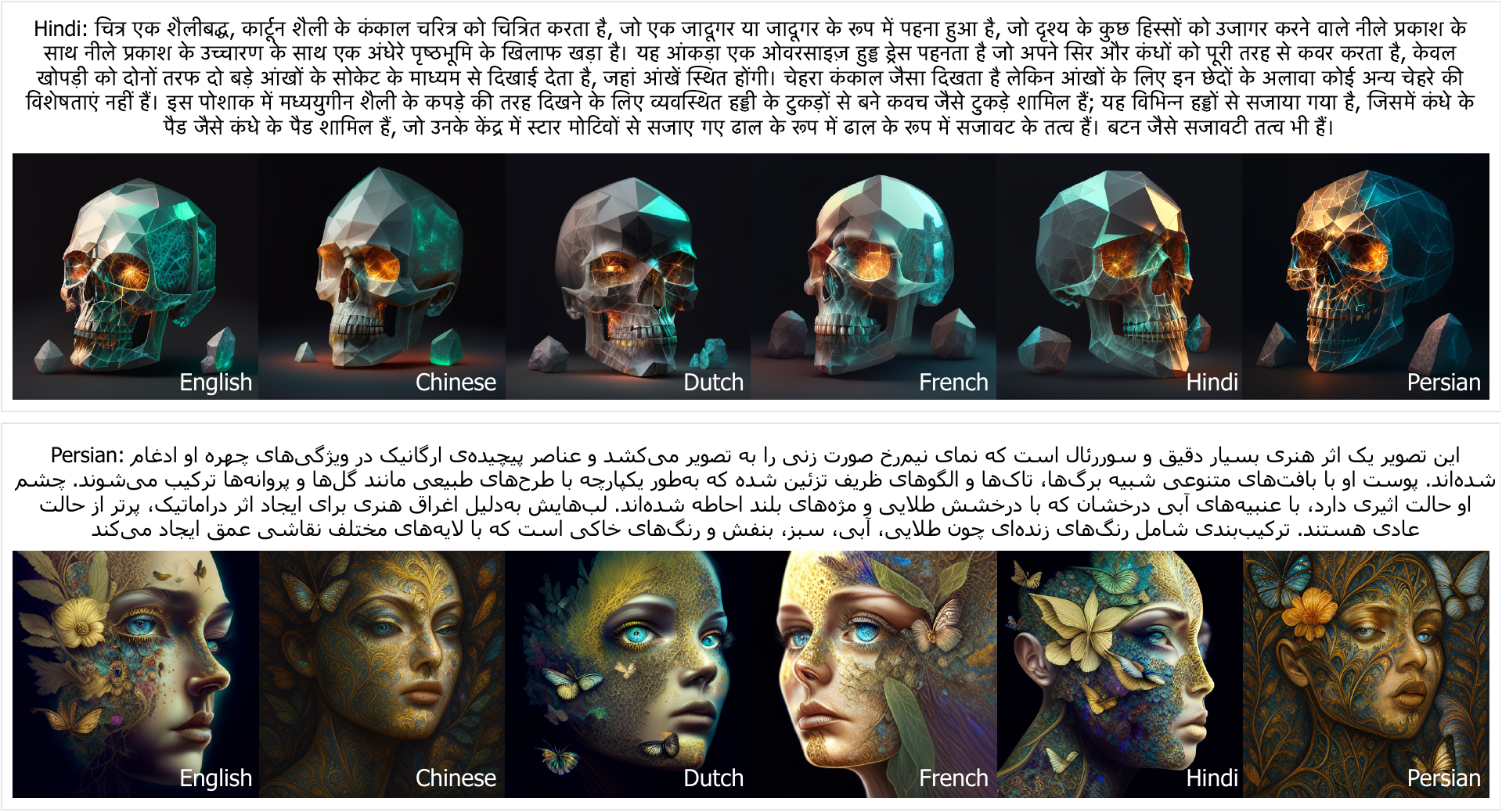}
    \caption{\textbf{Qualitative evaluation of \ours.} Each row is based on a single concept expressed in six different languages. For clarity, we show only one of the prompts (in one language) and present six images generated from its translated prompts in the other five languages. No matter the language, \ours consistently produces semantically aligned, visually coherent results.}
    \label{fig:qual-eval-2}
\end{figure}

To complement the quantitative findings, we present qualitative results from \ours across diverse prompt categories, including compositional scenes, abstract concepts, and multilingual instructions, in Figures~\ref{fig:qual-eval-1} and \ref{fig:qual-eval-2} (main paper) and Figure~\ref{fig:qual-eval-3} (appendix). The results show that \ours consistently generates semantically aligned and visually coherent images. Objects, layouts, and attributes are preserved across languages, demonstrating the model’s strong multilingual alignment and consistency in representing concepts.
% \begin{figure}[h]
%     \centering
%     \includegraphics[width=0.85\linewidth]{assets/qual_eval_1.png}
%     \caption{TBD}
%     \label{fig:qual-eval-2}
% \end{figure}

\begin{figure}[th!]
    \centering
\includegraphics[width=\linewidth]{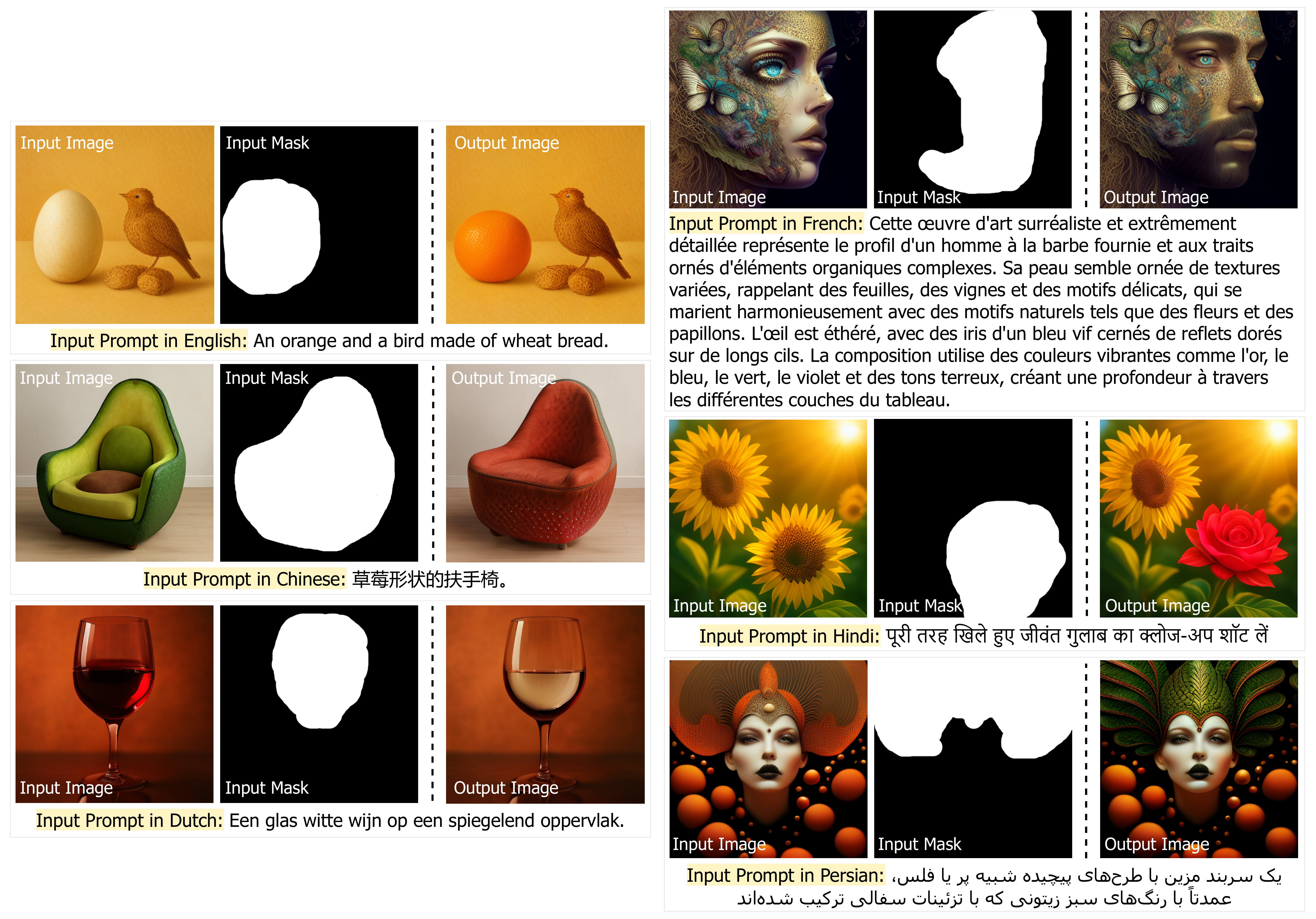}
    \caption{\textbf{Multilingual image inpainting}. \ours supports multilingual text-guided image inpainting, highlighting its potential for interactive and language-inclusive visual editing across diverse user groups.}
    \label{fig:inpainting}
\end{figure}

\begin{figure}[th!]
    \centering
\includegraphics[width=1\linewidth]{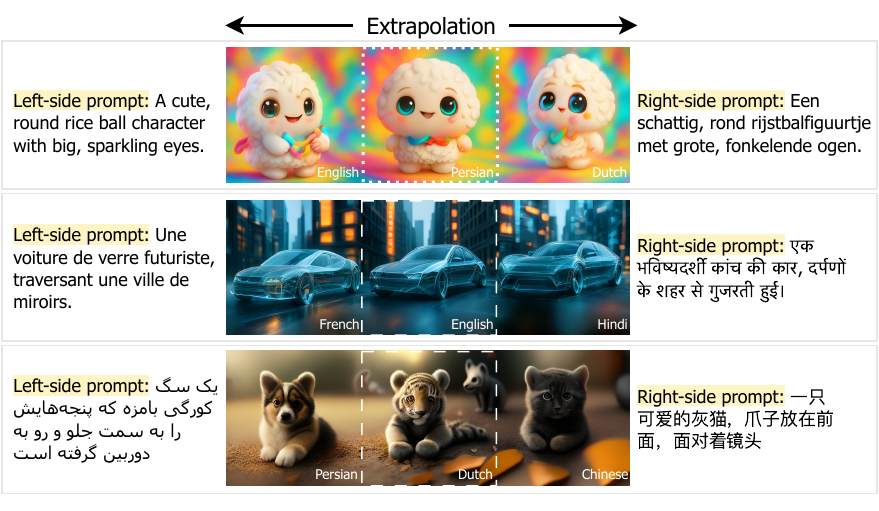}
    \caption{\textbf{Multilingual image extrapolation.} \ours successfully performs text-guided image extrapolation using multilingual prompts. Given the middle image and two different multilingual prompts (for the left and right extensions), \ours generates coherent visual completions on both sides, demonstrating extrapolation capability.}
    \label{fig:extrapolation}
\end{figure}

\subsubsection{Multilingual Image Inpainting and Extrapolation}
\ours enables new collaborative applications, such as a multilingual visual canvas where users can contribute prompts in their native languages to co-create coherent and expressive visual scenes. It supports text-guided image inpainting and extrapolation across multiple languages without requiring additional fine-tuning. As shown in Figures~\ref{fig:inpainting} and \ref{fig:extrapolation}, \ours can modify or extend an input image based on prompts in different languages, producing results that remain semantically faithful and visually consistent with the adjacent visual content. These examples demonstrate the model’s ability to maintain coherence across inpainted and extrapolated regions, highlighting its potential for interactive and multilingual visual editing.

\subsubsection{Cross-lingual Image Generation} 
\begin{figure}[th!]
    \centering
    \includegraphics[width=\linewidth]{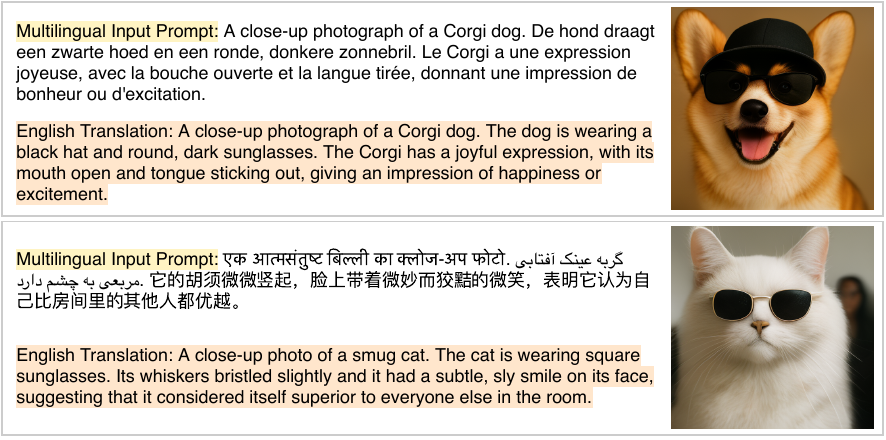}
    \caption{\textbf{Cross-Lingual Prompt Generation.} Examples of code-switched prompts mixing three languages, along with images generated by \ours. Top: English, Dutch and French. Bottom: Hindi, Persian and Chinese. English translations are shown below each prompt for reader convenience, they are not used as input.}
    \label{fig:cross-lingual-qualitative}
\end{figure}

A more challenging evaluation of the model’s multilingual ability involves prompts that combine multiple languages within the same input. This requires the model to integrate information from different languages into a coherent and semantically accurate image. To create these prompts, we split a base prompt into three parts and translate each into a different language. Figure~\ref{fig:cross-lingual-qualitative} (in the Introduction) illustrates two such examples. The images generated by \ours demonstrate its ability to follow complex multilingual instructions, producing visually coherent and semantically faithful outputs. These results highlight the model’s cross-lingual alignment, despite not being explicitly trained for this task.

\section{Ablations and Analyses}
\label{sec:ablation}

In the following sections, we conduct a series of ablation studies and analyses to evaluate the effects of progressive pretraining (Section~\ref{abl:progressive_pretraining}), instruction tuning (Section~\ref{abl:instruction_tuning}), and model merging (Section~\ref{abl:model-merging}). We also perform additional multilingual evaluations, including cross-linguistic consistency (Section~\ref{abl:clc}) and code switching similarity analyses (Section~\ref{abl:css}).

\subsection{Effect of Progressive Pretraining}
\label{abl:progressive_pretraining}

\begin{figure}[t]
    \centering
    \includegraphics[width=\linewidth]{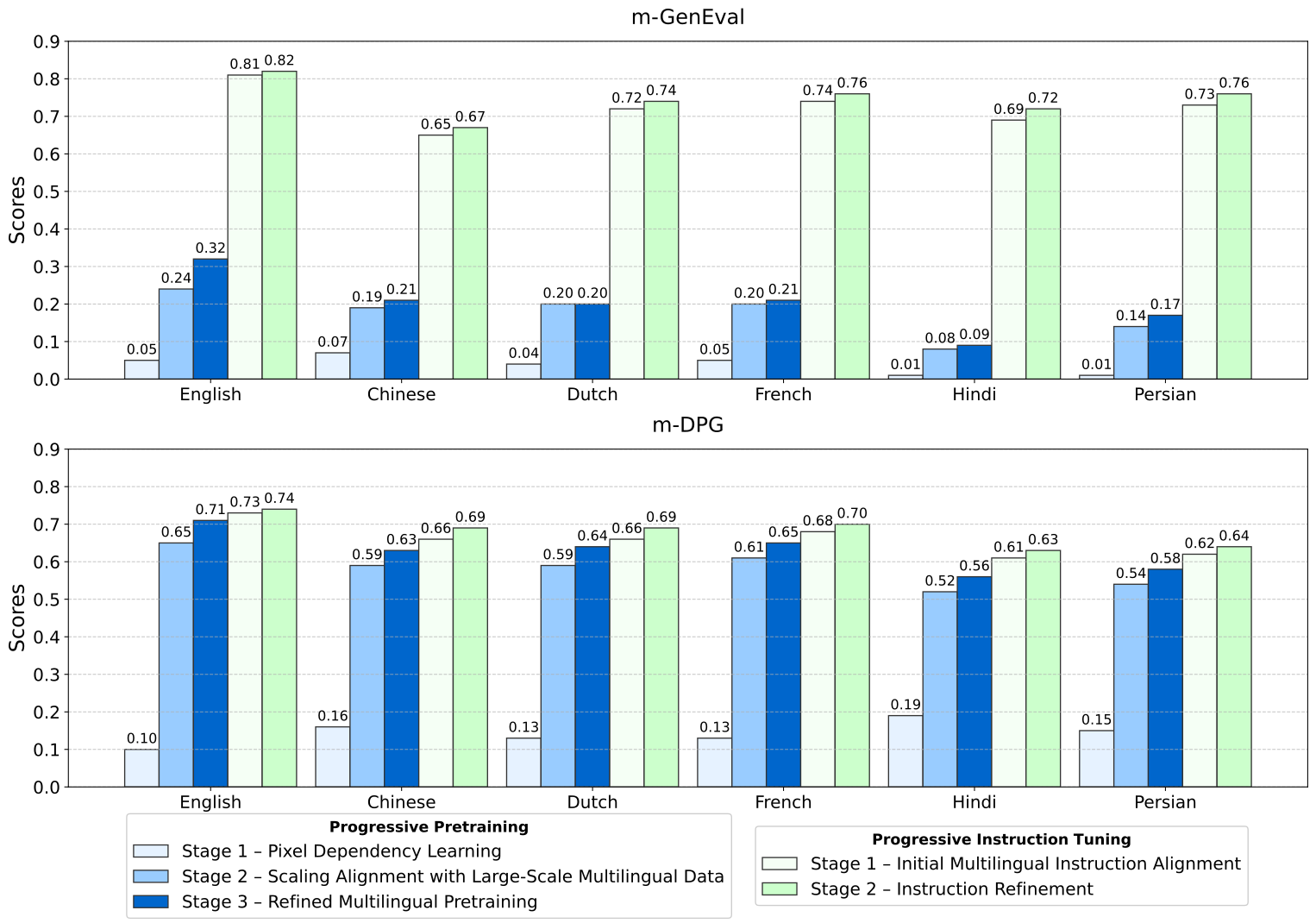}
    \caption{\textbf{Effect of Progressive Pretraining and Instruction Tuning.} Performance on m-GenEval (top) and m-DPG (bottom) improves steadily across pretraining and instruction tuning stages. Pretraining at $256{\times}256$ yields significant gains—especially on m-DPG—when scaling to large multilingual datasets (Stage 2), followed by refinement using higher-quality data (Stage 3). Instruction tuning at $512{\times}512$ brings a substantial boost (e.g., +0.52 on m-GenEval) and further improvement by increasing the share of curated, instruction-aligned samples. These results highlight how scale drives broad generalization, while data quality and resolution are key for performance refinement.}

    \label{fig:progressive-pretraning}
\end{figure}

We first analyze the impact of our progressive pretraining strategy across three stages at $256{\times}256$ resolution. As shown in Figure~\ref{fig:progressive-pretraning}, each stage leads to steady improvements in multilingual performance on m-GenEval and m-DPG. In the first stage, using only m-ImageNet 1K, the average scores are modest: 0.04 on m-GenEval and 0.14 on m-DPG, indicating weak multilingual alignment. In the second stage, the addition of large-scale but noisy datasets (m-SA-1B, m-CC12M, m-LAION-Aesthetic) results in a significant increase, reaching 0.17 on m-GenEval and 0.58 on m-DPG. The average gain in performance from stage one to stage two is substantially larger on m-DPG (0.44) than on m-GenEval (0.14), suggesting that this stage improves the model's ability to handle natural, descriptive prompts in multiple languages. The third stage incorporates higher-quality datasets (m-LAION-Aesthetic and m-JourneyDB), leading to further gains: 0.02 on m-GenEval and 0.04 on m-DPG. These results demonstrate the cumulative benefits of progressively increasing both the diversity and quality of pretraining data. While large, noisy datasets drive early generalization, high-quality data is essential for refining multilingual alignment. This staged pretraining approach provides a strong initialization for downstream instruction tuning.

% english, dutch, french, chinese, hindi, persian

\subsection{Effect of Progressive Instruction Tuning}
\label{abl:instruction_tuning}
We analyze the effect of high-resolution instruction tuning using a fixed dataset mixture: m-LAION-Aesthetic, m-JourneyDB, and m-BLIP3o-Instruct, all at $512{\times}512$ resolution. As shown in Figure~\ref{fig:progressive-pretraning}, both tuning stages progressively refine multilingual alignment, with consistent gains across all six languages. In the first stage, each training batch consists of 60\% m-LAION-Aesthetic, 30\% m-JourneyDB, and 10\% m-BLIP3o-Instruct samples. This stage yields a substantial multilingual gain of 0.52 on m-GenEval and 0.04 on m-DPG compared to the final stage of pretraining, indicating that high-resolution supervision provides strong improvements when combined with highly curated instruction tuning dataset.
In the second stage, each training batch shifts emphasis toward higher-quality and instruction-aligned data, consisting of 25\% m-LAION-Aesthetic, 60\% m-JourneyDB, and 15\% m-BLIP3o-Instruct samples. This leads to a further multilingual gain of 0.02 in m-GenEval and a boost in m-DPG. These results show that beyond increasing the resolution, the relative weight of curated and instruction-focused datasets plays a pivotal role in shaping multilingual capability. Prioritizing high-quality supervision at higher resolution proves effective for achieving competitive alignment ultimately enabling our 2B model to rival much larger models.

\subsection{Effect of Model Merging on Generalization}
\label{abl:model-merging}

\begin{wraptable}{r}{9cm} % [r]ight or [l]eft alignment, {width}
\centering
\small
\begin{tabular}{lcc}
\toprule
\textbf{Method} & \textbf{m-GenEval} & \textbf{m-DPG} \\
\midrule
Last checkpoint & 0.81  & 0.73 \\
\hdashline
EMA             & 0.82  & 0.75 \\
WMA             & 0.83  & 0.75 \\
SMA             & \textbf{0.83}  & \textbf{0.75} \\
\bottomrule
\end{tabular}
\caption{\textbf{Effect of model merging on generalization.} Without any fine-tuning, all merging strategies improve performance on m-GenEval and slightly enhance m-DPG, highlighting model merging as a simple yet effective way to boost generalization. This ablation uses English prompts.}
\label{tab:model_merging}
\end{wraptable}

We investigate the impact of model merging on multilingual image generation performance by combining $N=20$ checkpoints sampled at 10,000-step intervals from the second instruction tuning stage (steps 0–200K). Table~\ref{tab:model_merging} reports the results of three merging strategies: Simple Moving Average (SMA), Exponential Moving Average (EMA), and Weighted Moving Average (WMA) compared to the last checkpoint baseline. As reported, m-GenEval score on English prompt improves from 0.81 to 0.83 after model merging. Both WMA and SMA reach this upper bound, indicating that merging checkpoints along the optimization path enhances semantic alignment. Moreover, m-DPG score on English prompt remains stable or show modest gains, suggesting that merging preserves the model's ability to accurately follow dense, attribute-rich prompts without sacrificing fine-grained multilingual grounding. Among the merging strategies, SMA performs best overall due to its uniform averaging over well-aligned checkpoints. EMA also improves results but remains more susceptible to short-term training noise. WMA offers a compromise by emphasizing later checkpoints, trading off stability for adaptability. These findings underscore that checkpoint merging can meaningfully enhance both compositional understanding and multilingual robustness, with SMA offering a simple yet effective strategy.

\subsection{Cross-Lingual Consistency Analysis}
\label{abl:clc}

This ablation examines how different models preserve consistency in image generation across languages, evaluated using the CLC score introduced in Section \ref{sec:multilingual_eval_metrics} with both EVA-CLIP and DINOv2 vision encoders (Table~\ref{tab:clc_scores}). 

\vskip -0.1in
\begin{wraptable}{r}{9cm}
\centering
\small
\begin{tabular}{lrcc}
\toprule
\textbf{Model} & \textbf{Params} & \textbf{EVA-CLIP} & \textbf{DINOv2} \\
\midrule
Show-o & 1.3B      & 0.47 & 0.16 \\
Janus & 1.3B       & 0.67 & 0.26 \\
Janus Pro & 7B     & 0.67 & 0.30 \\
BLIP3-o & 4B       & 0.76 & 0.44 \\
BLIP3-o & 8B       & 0.77 & 0.45 \\
\midrule
\rowcolor{LightBlue} \ours & 2B      & \textbf{0.79} & \textbf{0.61} \\
\bottomrule
\end{tabular}
\caption{\textbf{Cross-Lingual Consistency Analysis} using CLC scores with EVA-CLIP and DINOv2 backbones. \ours (2B) outperforms larger models, showing stronger cross-lingual consistency in both semantic and visual domains. The larger gap in DINOv2 scores highlights its greater sensitivity to visual-structural variations, revealing inconsistencies that EVA-CLIP's semantic focus may overlook.}
\label{tab:clc_scores}
\end{wraptable}

Across both backbones, \ours achieves the highest scores, 0.79 (EVA-CLIP) and 0.61 (DINOv2), outperforming larger models such as BLIP3-o 8B (0.77/0.45) and Janus Pro 7B (0.67/0.30). 

This indicates that training strategy and data alignment play a more significant role than parameter count alone in achieving cross-lingual consistency.
In the EVA-CLIP based CLC, which emphasizes high-level semantic similarity due to its contrastive training objective with text, high scores reflect strong cross-lingual consistency in scene-level concept. In contrast, DINOv2-based CLC captures visual-structural coherence making it more sensitive to differences in object composition, layout, or fine-grained visual patterns. Notably, the relative margin between \ours and other models is more pronounced under DINOv2. For instance, BLIP3-o's DINOv2 score drops significantly despite competitive EVA-CLIP CLC score suggesting its outputs vary more in visual structure across languages. 

\begin{wraptable}{r}{9cm}
\centering
\small
\begin{tabular}{l r cc cc}
\toprule
 &  & \multicolumn{2}{c}{\textbf{EVA-CLIP}} & \multicolumn{2}{c}{\textbf{DINOv2}} \\
 \cmidrule(lr){3-4} \cmidrule(lr){5-6}
 \textbf{Model} & \textbf{Params} & \textbf{EF} & \textbf{ES} & \textbf{EF} & \textbf{ES} \\
\midrule
Show-o        & 1.3B                 & 0.73        & 0.72                & 0.41        & 0.38 \\
Janus         & 1.3B                 & 0.75        & 0.73                & 0.50        & 0.43 \\
Janus Pro     & 7B                   & 0.76        & 0.72                & 0.58        & 0.50 \\
BLIP3-o       & 4B                   & 0.75        & 0.75                & 0.54        & 0.54 \\
BLIP3-o       & 8B                   & 0.74        & 0.74                & 0.52        & 0.51 \\
\midrule
\rowcolor{LightBlue} \ours      & 2B                   & \textbf{0.82} & \textbf{0.81}      & \textbf{0.67} & \textbf{0.64} \\
\bottomrule
\end{tabular}
\caption{\textbf{Code Switching Similarity (CSS) Analysis} using EVA-CLIP and DINOv2 backbones. Scores are reported for two prompt variants: English First (EF) and English Second (ES). \ours (2B) outperforms larger models, showing strong visual consistency and robustness to code-mixed input order. The larger DINOv2 gap reflects its higher sensitivity to visual-structural variation, while EVA-CLIP remains more stable due to its semantic focus.}
\label{tab:cms_scores_combined}
\end{wraptable}

\begin{figure}[t!]
    \centering
    \begin{minipage}[t]{0.48\textwidth}
        \centering
        \includegraphics[width=\linewidth]{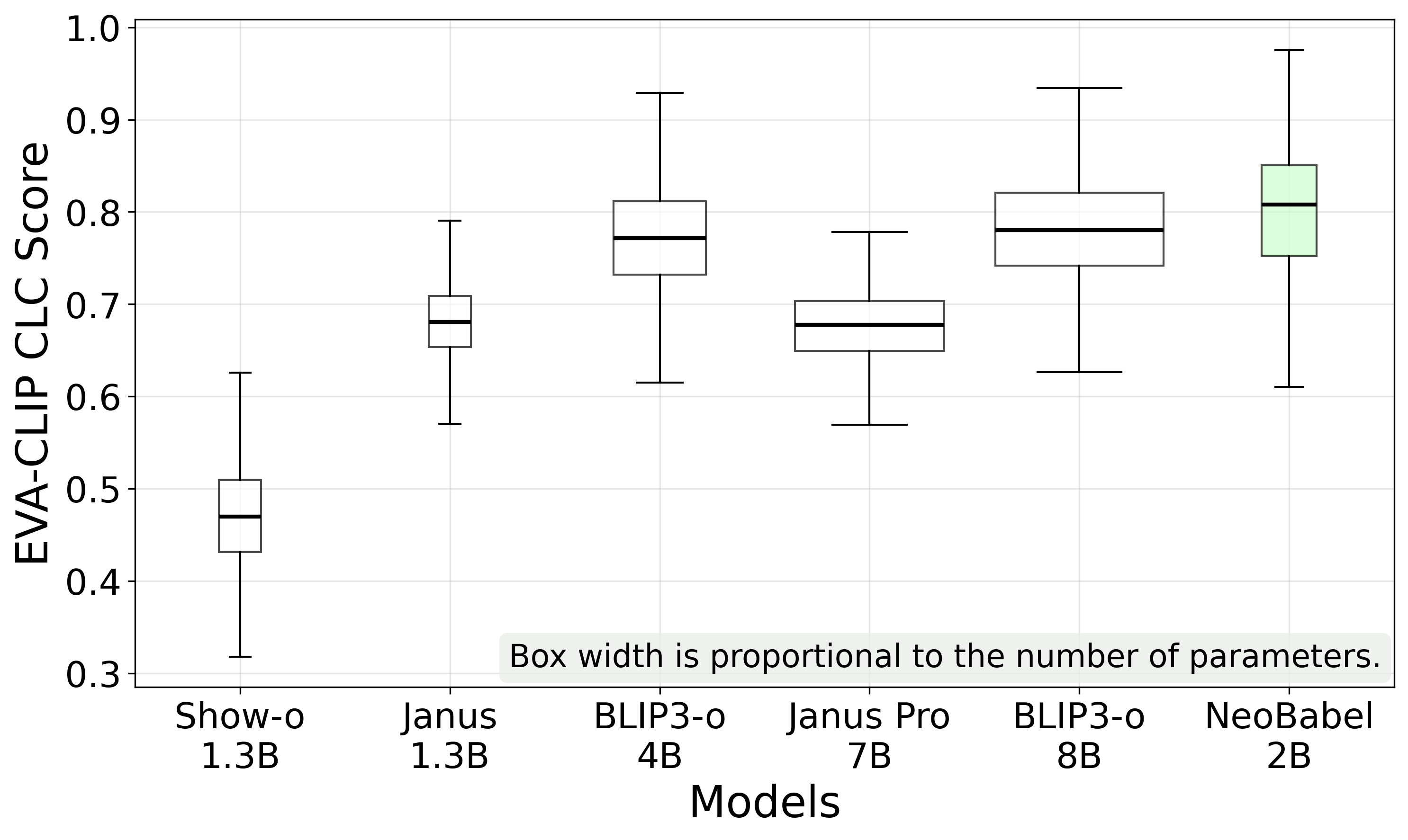}
        % \captionof{figure}{Caption for Image 1}
        % \label{fig:image1}
    \end{minipage}
    % \hfill
    \begin{minipage}[t]{0.48\textwidth}
        \centering
        \includegraphics[width=\linewidth]{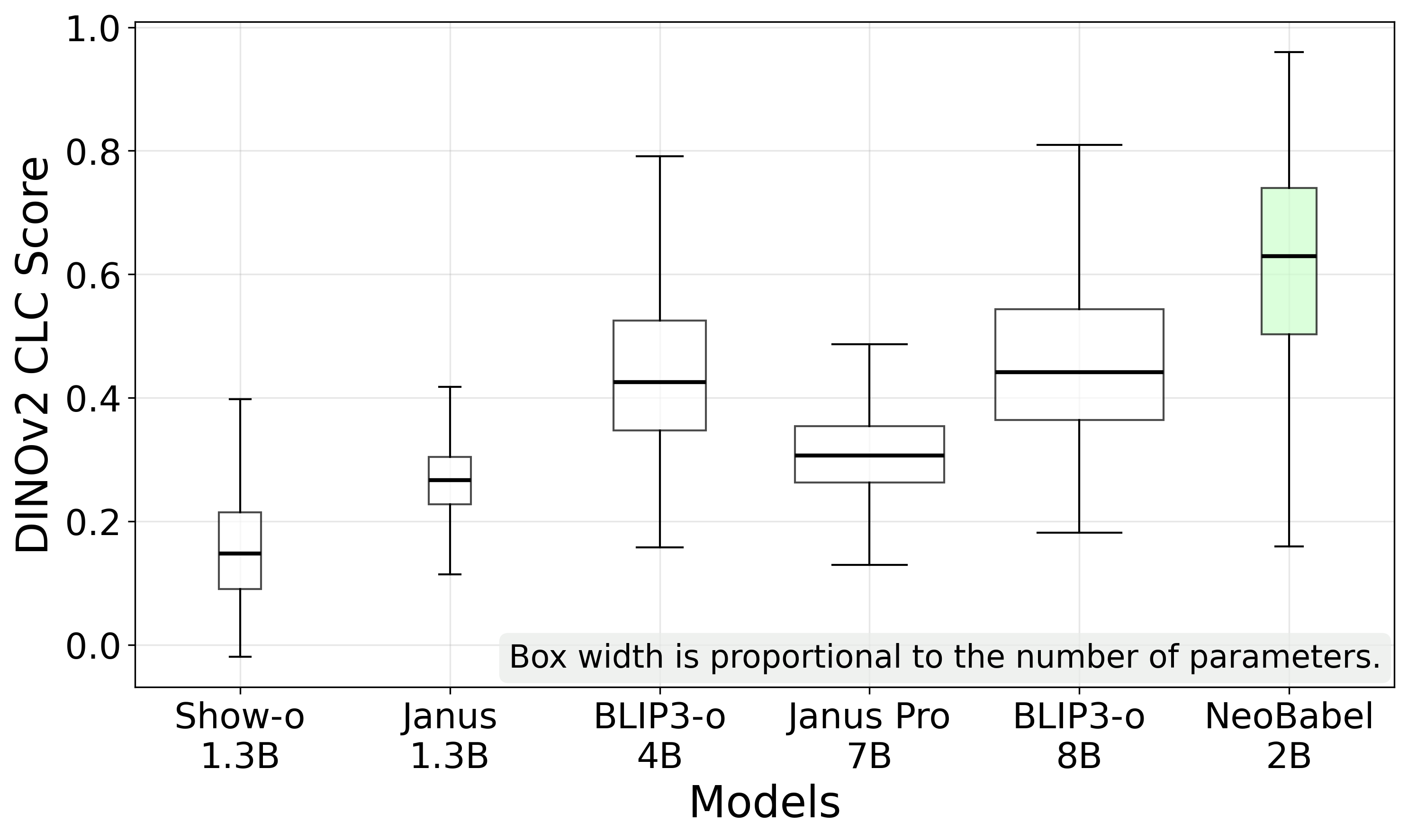}
        % \captionof{figure}{Caption for Image 2}
        % \label{fig:image2}
    \end{minipage}
    \caption{\textbf{Cross-Lingual Consistency (CLC) Score Distributions across Models.} 
We show the distribution of CLC scores computed using EVA-CLIP (left column) and DINOv2 (right column), where higher values reflect greater consistency across languages. EVA-CLIP captures semantic similarity, while DINOv2 is more sensitive to visual structure and layout. \ours achieves the highest scores under both backbones—particularly with DINOv2—demonstrating strong alignment in both meaning and visual form. Box widths reflect model size, showing that bigger models aren’t always more consistent.}
    \label{fig:variation_clc}
\end{figure}

This amplifies the role of multilingual alignment not just in semantics, but in visual form a dimension better captured by DINOv2. \ours achieves high scores under both backbones, thereby demonstrating strong cross-lingual consistency in both scene semantics and visual structure. Figure~\ref{fig:variation_clc} complements Table~\ref{tab:clc_scores} by visualizing the distribution of CLC scores for each model using both EVA-CLIP and DINOv2 backbones. In terms of EVA-CLIP based CLC variation, \ours performs on par with BLIP3-o 8B, the second-best model. However, under the DINOv2-based CLC variation, \ours outperforms all baselines, exhibiting lower dispersion and higher consistency. We can also observe from the figure that simply increasing model size does not guarantee better cross-lingual consistency, as evidenced by the lower DINOv2 scores of Janus Pro 7B and BLIP3-o 8B compared to \ours.

% \subsection{Crosslingual Semantic Drift Coefficient}

% Question: How does visual meaning drift across cultural-linguistic boundaries?

\subsection{Code Switching Similarity Analysis}
\label{abl:css}

\begin{figure}[ht]
    \centering

    % First Row: EVA-CLIP
    \begin{subfigure}[t]{0.48\textwidth}
        \centering
        \includegraphics[width=\linewidth]{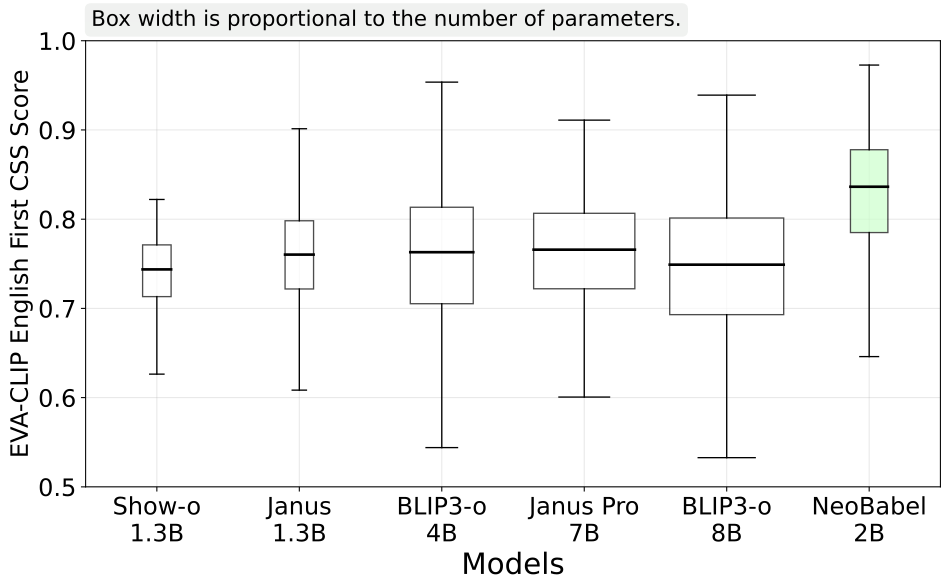}
    \end{subfigure}
    \hfill
    \begin{subfigure}[t]{0.48\textwidth}
        \centering
        \includegraphics[width=\linewidth]{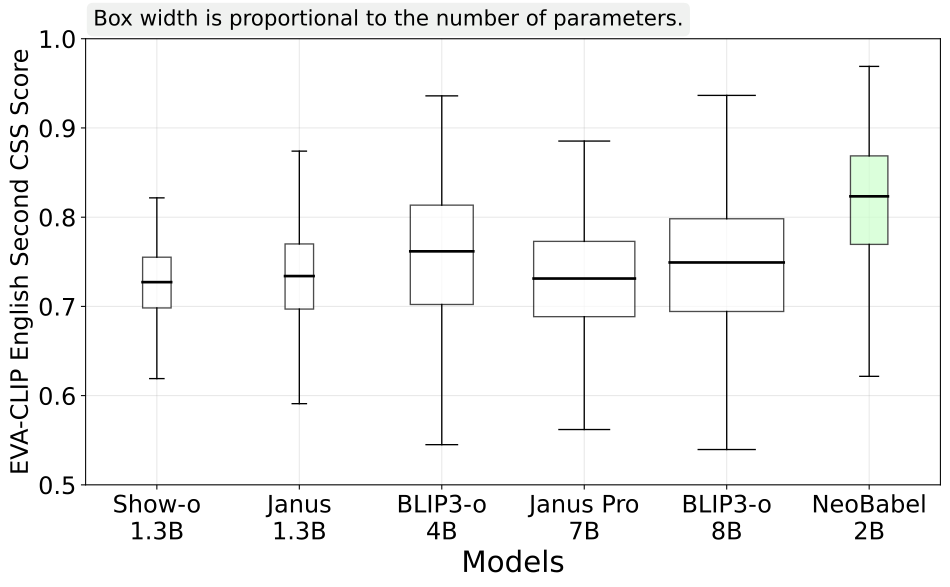}
    \end{subfigure}

    \vspace{0.5em}

    % Second Row: DINOv2
    \begin{subfigure}[t]{0.48\textwidth}
        \centering
        \includegraphics[width=\linewidth]{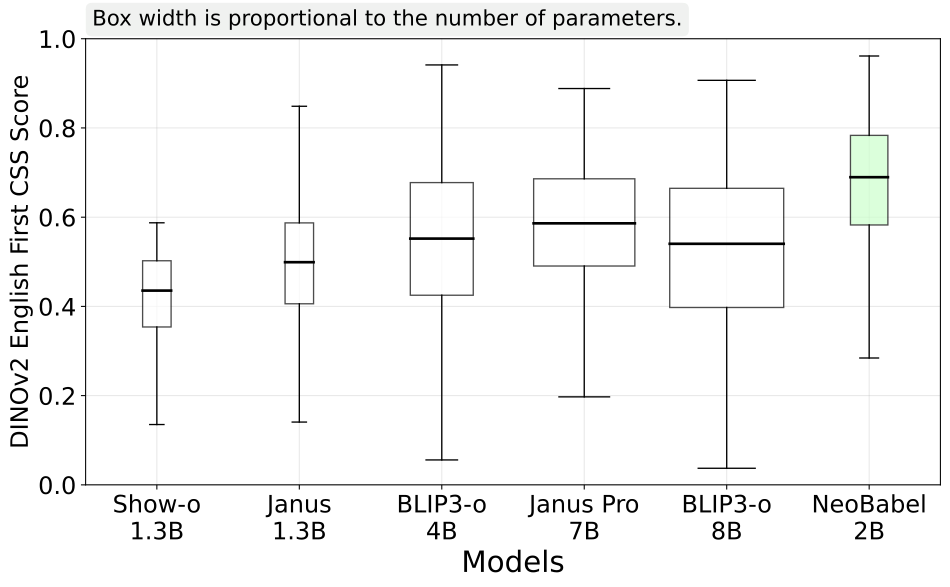}
    \end{subfigure}
    \hfill
    \begin{subfigure}[t]{0.48\textwidth}
        \centering
        \includegraphics[width=\linewidth]{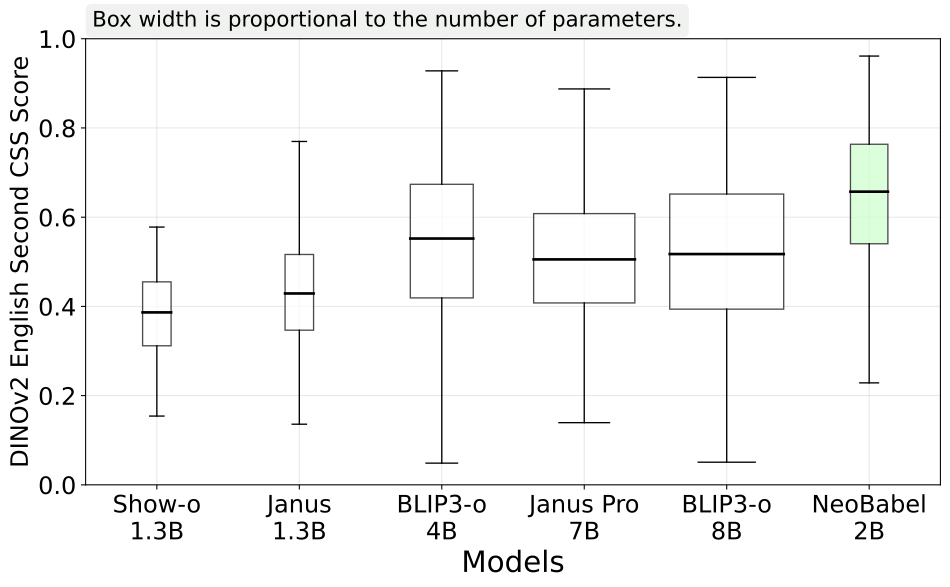}
    \end{subfigure}

    \caption{
    \textbf{Variation in Code Switching Similarity (CSS) Scores across Models.} We report CSS scores for code-mixed prompts under two settings: English-first (left column) and English-second (right column), using EVA-CLIP (top row) and DINOv2 (bottom row) as backbones. Higher scores indicate stronger visual alignment with the reference image, while smaller EF–ES gaps suggest robustness to code-switch position. \ours consistently achieves higher medians and lower variance than larger baselines, especially under DINOv2, highlighting its effective and stable handling of multilingual prompts.}

    \label{fig:code_switching_combined}
\end{figure}

We evaluate model robustness to intra-prompt code switching using the proposed CSS score in Section \ref{sec:multilingual_eval_metrics} with EVA-CLIP and DINOv2 backbones. As shown in Table~\ref{tab:cms_scores_combined} and Figure~\ref{fig:code_switching_combined}, \ours consistently outperforms larger models, demonstrating stronger visual consistency under both English-First (EF) and English-Second (ES) variations. Under EVA-CLIP, all models exhibit minimal difference between EF and ES, suggesting that the position of the English segment has limited effect on global semantic alignment. In contrast, DINOv2 scores are lower across the board, indicating greater difficulty in maintaining consistent visual structure when mixing languages. 

Importantly, a desirable outcome is both high CSS scores (indicating alignment with the reference image) and minimal gap between EF and ES (indicating robustness to code-switch position). \ours achieves this balance, with CSS scores of 0.82 (EF) and 0.81 (ES) for EVA-CLIP, and 0.67 (EF) and 0.64 (ES) for DINOv2. The box plots reveal that although larger models like BLIP3-o (8B) achieve competitive means, they show greater variability across prompts. \ours demonstrates both higher median performance and lower dispersion, confirming its consistent handling of code-mixed inputs. These results further highlight that scaling model size does not necessarily improve code-mixed prompt robustness effective multilingual alignment plays a larger role.

\section{Related Works}
\label{sec:related_work}

\textbf{Large Multimodal Models.}
Recent advances in large multimodal models (LMMs)~\citep{llava, internvl, llavaonevision, Qwen2.5-VL, dash2025ayavisionadvancingfrontier} have extended large language models (LLMs)~\citep{llama, qwen2.5} to support image understanding tasks, including image captioning and visual question answering. These models typically rely on a vision encoder to extract image features, which are then projected into the LLM embedding space for cross-modal alignment. More recent encoder-free models~\citep{xie2024show, diao2024EVE, diao2025EVEv2} bypass the explicit image encoder and instead align raw visual tokens directly within the LLM space. Among early efforts to enable multilingual visual understanding, Maya \citep{Maya, BehindMaya}, Aya-Vision~\citep{dash2025ayavisionadvancingfrontier} and Pangea~\citep{pangea} incorporate a multilingual training corpus. However, they are limited to image understanding tasks. In contrast, our proposed \ours architecture focuses exclusively on multilingual image generation, offering the first encoder-free model that aligns visual features in the LLM space while supporting cross-lingual generation. Architecturally, \ours is closely related to show-o~\citep{xie2024show}, sharing the same design goal of direct visual alignment in language space, but differs in its task focus and multilingual design.

\textbf{Visual Generative Models.} Two dominant paradigms have emerged for image and video generation: diffusion-based~\citep{rombach2022high, peebles2023scalable, uvit, pixart, Xie_2023_ICCV, wu2023tune, lipman2022flow, xie2025sana, qin2025lumina, show1, seawead2025seaweed} and autoregressive~\citep{sun2024autoregressive, kondratyuk2023videopoet, chen2020generative, randar, ARPG} models. Diffusion models typically combine pretrained text encoders with denoising networks to iteratively refine visual outputs, while autoregressive models adopt LLM-based architectures trained via next-token prediction. Recent hybrid approaches~\citep{mar, mardini, unifluid} attempt to unify the strengths of both paradigms for more powerful generation. \ours follows the diffusion-based paradigm but distinguishes itself by adopting an LLM-style architecture for visual token modeling. This removes the reliance on frozen text encoders and instead builds on top of a strong multilingual decoder-based LLM, enabling tighter integration between language and vision.

\textbf{Unified Multimodal Models.} Unified multimodal models aim to handle both image understanding and generation within a single architecture, typically categorized into native and adapter-based approaches.  Native approaches such as Chameleon~\citep{team2024chameleon}, Show-o~\citep{xie2024show}, and Transfusion~\citep{zhou2025transfusion} adopt either autoregressive, diffusion, or hybrid modeling strategies to jointly process vision and language. Recent work~\citep{wang2024emu3, vila-u, unitok, unitoken, SemHiTok, dualtoken} has focused on improving tokenization and training efficiency to enhance cross-modal alignment. A parallel direction~\citep{tang2023any, uio2, dreamllm, ge2024seed, tong2024metamorph, metaqueries, chen2025blip3, wu2023next} constructs unified multimodal models by connecting pretrained LMMs and generative models via adapters or learnable tokens. While modular and flexible, these systems often rely on frozen components and lack full cross-modal integration. Our model, \ours, aligns more closely with native unified multimodal models by unifying visual and textual modeling within a single decoder-based architecture, without relying on adapters or frozen backbones. Although \ours supports multilingual multimodal understanding, this work focuses specifically on multilingual image generation.

\section{Limitations}
\label{sec:limitation}
While \ours demonstrates strong multilingual image generation capabilities, several limitations remain. First, the model currently supports only six languages; extending to broader linguistic coverage would require further tokenizer adaptation and additional training. Second, although \ours adopts a unified architecture, it does not yet support vision-language tasks such as visual question answering, due to the absence of task-specific fine-tuning. Third, the model's performance is constrained by its parameter size and the diversity and quality of the training data. For instance, during instruction tuning, we used a fixed mixture of m-LAION-Aesthetic, m-JourneyDB, and m-BLIP3o-Instruct, without performing an extensive sweep over mixture ratios—an area that could reveal further improvements. We leave these directions, including task expansion, larger-scale scaling, and wider language coverage, for future research.

\section{Conclusion}
\label{sec:conclusion}

\ours demonstrates that high-quality, efficient multilingual image generation is not only possible but also advantageous. 
Through strategic data curation and a unified architecture, we set a new Pareto frontier in performance, efficiency, and inclusivity. While currently focused on text-to-image generation, our model is structurally capable of broader multimodal tasks. 
Our results across m-GenEval and m-DPG benchmarks, paired with the introduction of new evaluation metrics (CLC and CSS), establish a robust foundation for the next generation of multilingual generative models. 

Our work opens several promising avenues for future research. First, extending \ours to encompass a wider variety of languages, particularly those currently underrepresented in vision-language research, remains an important objective. The modularity of our framework and the accompanying open-source toolkit are designed to facilitate such extensions.
Second, beyond linguistic diversity, integrating cultural grounding into multimodal models presents a compelling research direction. By curating datasets annotated with region-specific concepts, aesthetic preferences, and social norms, future work could develop models that are not only multilingual but also culturally aware and adaptive.
Finally, this work contributes to the broader goal of democratizing generative AI. By releasing all model weights, datasets, and evaluation protocols, we aim to encourage the research community to build upon this foundation, ultimately advancing toward generative models that better reflect and serve global linguistic and cultural diversity.

\section*{Acknowledgment}
We would like to thank the Cohere Labs team for their valuable feedback and for providing generous computing resources for conducting and analyzing our experiments.
We further acknowledge the Dutch Research Council (NWO) in The Netherlands for awarding this project access to the LUMI
supercomputer, owned by the EuroHPC Joint Undertaking, hosted by CSC (Finland) and the LUMI consortium through
the Computing Time on National Computer Facilities call. 
We also acknowledge NWO for providing access to Snellius, hosted by SURF through the Computing Time on National Computer Facilities call for proposals.
Cees G. M. Snoek is (partially) funded by the Horizon Europe project ELLIOT (GA No. 101214398).

\vspace{2em}
\bibliography{references}
\newpage
\section*{Appendix A}

\begin{figure}[th!]
    \centering
    \includegraphics[width=\linewidth]{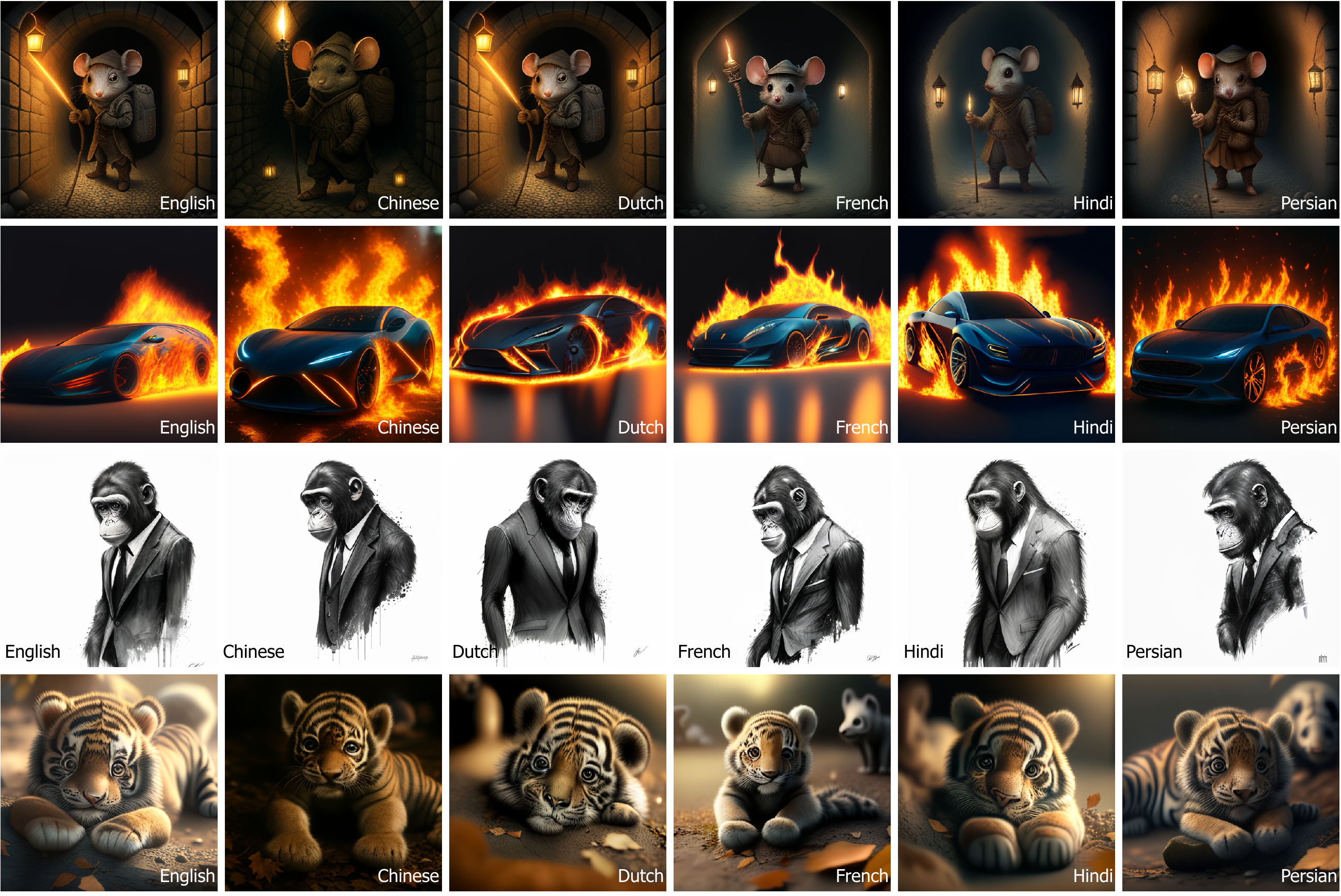}
    \caption{\textbf{Qualitative Evaluation of \ours.} 
Each row corresponds to a single concept expressed in six different languages: English, Chinese, Dutch, French, Hindi, and Persian. 
Although prompts are not shown for readability, all images were generated using translated versions of the same underlying prompt in each language. 
\ours consistently produces semantically aligned and visually coherent results across languages, highlighting its strong multilingual generation capabilities. We intentionally omit the prompts here due to their length, focusing instead on the visual consistency across languages.}
\label{fig:qual-eval-3}
\end{figure}

This appendix provides additional training details and qualitative results to supplement the main paper. Table~\ref{tab:hyperparameters} outlines the key hyperparameters used across the three pretraining stages and two instruction tuning stages of \ours. Figure~\ref{fig:qual-eval-3} presents representative multilingual generation examples, demonstrating visual consistency across six languages.

\begin{table}[tbh!]
\centering
\arrayrulecolor{black}
\resizebox{\textwidth}{!}{%
\begin{tabular}{lccc|cc}
\toprule
  & \multicolumn{3}{c|}{\textbf{Pretraining}} 
  & \multicolumn{2}{c}{\textbf{Instruction Tuning}} \\
  % & \multicolumn{3}{c|}{} 
  % & \multicolumn{3}{c}{} \\
 \textbf{Hyperparameters}  & \textbf{1st Stage} & \textbf{2nd Stage} & \textbf{3rd Stage} 
  & \textbf{1st Stage} & \textbf{2nd Stage} \\
  %& \textit{ImageNet 1K} & \textit{CC12M, SAM1B, LAION12M} & \textit{LAION12M, JourneyDB} 
  %& \textit{ImageNet 1K} & \textit{CC12M, SAM1B, LAION12M} & \textit{LAION12M, JourneyDB} \\
\cmidrule(lr){1-4} \cmidrule(lr){5-6}
Training Steps            & $500{k}$ & $500{k}$ & $500{k}$ & $500{k}$ & $200{k}$  \\
Warmup Steps              & $5000$ & $5000$     & $5000$ & $5000$ & $2000$  \\
Learning Rate             & $1e-4$ & $1e-4$     & $1e-4$ & $2e-4$ & $5e-05$  \\
Learning Rate Decay       & cosine & cosine     & cosine & cosine & cosine  \\
Optimizer                 & AdamW & AdamW & AdamW & AdamW & AdamW  \\
Image Resolution          & $256 \times 256$ & $256 \times 256$ & $256 \times 256$ & $512 \times 512$ & $512 \times 512$  \\
LLM Sequence Length       & $128$ & $512$ & $512$ & $512$ & $512$  \\
LLM Vocab Size            & $256{k}$ & $256{k}$ & $256{k}$ & $256{k}$ & $256{k}$  \\
Codebook Size             & $8192$ & $8192$ & $8192$ & $8192$ & $8192$  \\
\bottomrule
\end{tabular}
}
\caption{\textbf{Hyperparameters across training progression.}}
\label{tab:hyperparameters}
\end{table}

\end{document}